%% file: neurips_2025.tex
\documentclass{article}

\PassOptionsToPackage{authoryear,round}{natbib}

\usepackage[preprint]{neurips_2025}

\usepackage{neurips_2025}

\usepackage[utf8]{inputenc} 
\usepackage[T1]{fontenc}    
\usepackage{hyperref}       
\usepackage{url}            
\usepackage{booktabs}       
\usepackage{amsfonts}       
\usepackage{nicefrac}       
\usepackage{microtype}      
\usepackage{xcolor}         
\input{commands}

\title{\method: Selective and Adaptive Retrieval-augmented Generation with Context Compression}

\author{Yiqiao Jin, Kartik Sharma  \\
Georgia Institute of Technology \\
\texttt{\{ksartik,yjin328\}@gatech.edu}
\And
Vineeth Rakesh, Yingtong Dou \\
Visa Research  \\
\texttt{\{vinmohan,yidou\}@visa.com} 
\AND
Menghai Pan, Mahashweta Das \\
Visa Research  \\
\texttt{\{menpan,mahdas\}@visa.com} 
\And
Srijan Kumar \\
Georgia Institute of Technology \\
\texttt{srijan@gatech.edu} 
}

\begin{document}

\maketitle

\input{src/abstract_vineeth}

\input{src/intro_vineeth}

\input{src/method_vineeth}

\input{src/eval}

\input{src/related}

\input{src/conclusion}


\bibliographystyle{plainnat}
\bibliography{citations}


\appendix

\input{src/appendix}

\end{document}

%% file: commands.tex
\usepackage{algorithm}
\usepackage{algorithmicx}
\usepackage{algpseudocode}
\usepackage{amsmath}
\usepackage{amsthm}
\usepackage{array}
\usepackage{booktabs}
\usepackage{calc}
\usepackage{colortbl}
\usepackage{enumitem}
\usepackage{footnote}
\usepackage{graphicx}
\usepackage{hyperref}
\usepackage{listings}
\usepackage{mathtools}
\usepackage{microtype}
\usepackage{multirow}
\usepackage{newfloat}
\usepackage{seqsplit}
\usepackage{subcaption}  
\usepackage{url}
\usepackage{xcolor}
\usepackage{xspace}
\usepackage{lineno}
\usepackage{wrapfig}
\usepackage{lipsum}
\usepackage{tabularx}

\newcommand{\black}[1]{\textcolor{black}{#1}}

\newcommand{\gray}[1]{\textcolor{gray}{#1}}

\newcommand{\method}{SARA\xspace}
\newcommand{\methodbf}{\textbf{SARA}\xspace}

%% file: src/abstract_vineeth.tex
\begin{abstract}
Retrieval‑augmented Generation (RAG) extends large language models (LLMs) with external knowledge but faces key challenges: restricted \emph{effective} context length and redundancy in retrieved documents. 
Pure compression-based approaches reduce input size but often discard fine-grained details essential for factual accuracy. 
We propose \method, a unified RAG framework that balances \emph{local precision} and \emph{global knowledge coverage} under tight context budgets. 
\method combines natural-language text snippets with semantic compression vectors to jointly enhance context efficiency and answer correctness. 
It represents contexts at two complementary levels: 1) \emph{fine-grained} natural-language spans that preserve critical entities and numerical values, and 2) \emph{compact}, \emph{interpretable} vectors that summarize high-level semantics. 
An iterative evidence-selection module employs the compression vectors for dynamic reranking of contexts. 
Across 9 datasets and 5 open-source LLMs spanning 3 model families (Mistral, Llama, and Gemma), \method consistently improves answer relevance (+17.71), answer correctness (+13.72), and semantic similarity (+15.53), demonstrating the importance of integrating textual and compressed representations for robust, context-efficient RAG. 
\end{abstract}

%% file: src/intro_vineeth.tex
\section{Introduction}
Large language models (LLMs) have demonstrated remarkable capabilities across various natural language understanding and generation tasks~\citep{xiao2023large,zhao2023competeai}. 
Meanwhile, as LLMs are parametric in nature, their knowledge is inherently constrained by the scope, domain, and recency of their training data
~\citep{jin2024better,liu2025culturevlm}. 
Retrieval-augmented generation (RAG)~\citep{lewis2020retrieval} addresses this by retrieving from external non-parametric knowledge sources, essential for knowledge-intensive tasks. 

\noindent \textbf{Challenges.} 
Despite its promise, RAG still faces key challenges in effectively retrieving, selecting, and integrating external evidence. 
\emph{1) Limited Effective Context.} 
While some LLMs support long inputs, 
their attention is biased toward earlier tokens~\citep{li2024understanding}, making them sensitive to input order and prone to overlooking important information near the end of the input~\citep{yurankrag}. 
Extending usable context often requires costly, model-specific architectural changes~\citep{ding2023longnet}. 
\emph{2) Context Redundancy.} 
Retrieved documents often include \emph{redundant} or loosely structured content (e.g. transcripts or news articles)~\citep{yurankrag, gecontext}. 
Without careful post-processing, duplicate or irrelevant content inflates token usage, distracts the model, degrades answer quality or even leads to hallucinations. 
\emph{3) Compression-fidelity Trade-off.} 
Existing context compression techniques reduces input length but often sacrifice fine-grained details (e.g.  numeric values, organization names, and geographical locations), leading to hallucinated or incomplete responses. 
While existing methods achieve high compression rates, aggressive compression process risk discarding critical information essential for factual accuracy. 

\noindent \textbf{This Work.}
We present \method, a unified RAG framework that improves both \emph{retrieval} and \emph{generation} stages through structured evidence compression and adaptive selection. 
From the \emph{generation} perspective, \method represents long contexts using a small number of semantically rich, self-contained \emph{compression vectors}, which act as lightweight abstractive summaries that preserve essential information while significantly reducing input length. Specifically, we leverage state-of-the-art embedding models~\citep{SFRAIResearch2024,muennighoff2023mteb} to encode retrieved documents into multiple, semantically rich compression vectors. 
These vectors are also \emph{explainable} and can be interpreted through auto-encoding to reveal their underlying semantics. 
From the \emph{retrieval} perspective, \method introduces an \emph{iterative evidence selection mechanism} that leverages the compression vectors to dynamically refine the set of top-ranked documents. \method progressively selects contexts based on the knowledge required to properly address the query and knowledge coverage of existing contexts, minimizing redundancy while maximizing informativeness. 
\method is agnostic to the choice of embedding models, open-source LLMs, and retrievers. Our contributions are as follows:
\vspace{-3mm}
\begin{itemize}[leftmargin=1em]
    \item We propose \method, a novel RAG framework for long-context tasks. 
    \method introduces a \textbf{hybrid compression strategy}, balancing \emph{local precision} using natural language spans and \emph{global abstraction} via compression vectors, enabling fine-grained reasoning and holistic understanding within strict context budgets. 
    \vspace{-2mm}
    \item We propose an \textbf{iterative context refinement} mechanism based on the compression vectors to dynamically optimize the retrieved context by reducing redundancy and prioritizing query-relevant content. 
    \vspace{-2mm}
    \item Comprehensive experiments on $5$ LLMs spanning $3$ model families, including Mistral-7B, MistralNemo-12B, MistralSmall-24B, Llama-3.1-8B, and Gemma3-4B, demonstrate that \method consistently improves performance and generalizes well across LLMs (Section~\ref{app:generalization_llms}), retrievers (Section~\ref{app:generalization_retrievers}), and embedding models (Appendix~\ref{app:generalization_embed}).   
\end{itemize}

%% file: src/method_vineeth.tex
\vspace{-2mm}
\section{Method}
\vspace{-2mm}
\subsection{Problem Formulation}
\vspace{-2mm}
A retrieval-augmented generation (RAG) pipeline consists of a \emph{retriever} that fetches relevant evidence from a large-scale corpus based on the input query 
and a \emph{generator} that 
synthesizes the evidence to answer the query. 
Given a query $q$ and corpus $C$, the retriever $\mathcal{R}(\cdot)$ selects the top-$n$ relevant contexts $\mathcal{V}_{\mathrm{sel}} \subseteq C$.  
To improve effectiveness, RAG may incorporate a \emph{reranking} step to reorder the input documents, prioritizing the most relevant ones for answer generation. 

\vspace{-1mm}
\subsection{Overview} 
\vspace{-1mm}
LLMs have limited effective context windows, and performance degrades when key information is buried in long inputs~\citep{jiang2023longllmlingua}. 
\method mitigates this by compressing long context into compact vectors while selectively retaining essential evidence in natural language, preserving model capacity for the most relevant content.

\method follows a two-stage training procedure:
During \textbf{Compression Learning}, \method learns to reconstruct original context from compression vectors.
In \textbf{Instruction-tuning}, \method is adapted to rerank the evidence using the compression vectors and reason over mixed inputs--combining natural language and compressed evidence.
Our method is \emph{model-agnostic}, compatible with any retrievers, embedding models, and open-source LLMs. 
A lightweight \emph{projection layer} aligns the embedding space with the LLM space, requiring no significant changes to internal components like the attention mechanism, enabling seamless integration with future embedding models and LLMs. 
Sample prompts for all stages are provided in Table~\ref{tab:prompt}.

\vspace{-1mm}
\subsection{Compression Learning}
\label{sec:compression_learning}
\vspace{-1mm}
\begin{figure*}[htbp]
\centering
\includegraphics[width=0.98\linewidth]{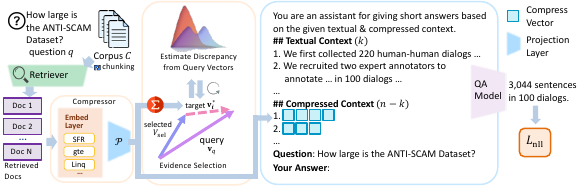}
\vspace{-1mm}
\caption{\method reasons over a mixture of compressed evidence and natural language contexts to balance local precision and global coverage when generating responses. An iterative evidence reranking step selects contexts for relevance and diversity. 
The retriever, compressor, and QA model uses a variety of embedding models. 
}
\label{fig:model}
\vspace{-2mm}
\end{figure*}

An effective compression mechanism should meet three core principles: 1) \emph{Semantic Fidelity}--preserving sufficient information for accurate context reconstruction; 2) \emph{Token Compatibility}--producing compression vectors interpretable by LLMs via prompting; and 3) \emph{Scalability}--
requiring minimal adaptation across retrievers and LLMs. 

To meet these goals, \method leverages sentence embeddings~\citep{reimers-2019-sentence-bert} aligned with the LLM's token space, enabling compact and interpretable representations that support reasoning under tight context budgets.

\noindent \textbf{Embedding Alignment.} 
\method encodes each text chunk into a compression vector that fits within a single token's embedding space (Figure~\ref{fig:model_pretrain}). 
A lightweight compressor--combining a sentence embedding model and an MLP--is trained via an autoencoding task~\citep{liu2023visual, cheng2024xrag} to align sentence embeddings with the LLM's token space:
\begin{equation}
\mathcal{L}_{\mathrm{align}} (s_i) = - \log P_{\theta}\left(s_i \mid \mathrm{Enc}(s_i), x_{\text {ins}}\right).
\label{eq:embed_align}
\end{equation}
Here, $s_i$ is a text chunk, $\mathrm{Enc}(\cdot)$ is the compressor, $\theta$ is the model's parameter, and $x_{\text {ins}}$ is the decoding instruction such as ``The token \texttt{<C>} can be interpreted as: \texttt{[CHUNK]}.'' 
As one compression vector has limited representation capacity, we segment each document into chunks, and encode each chunk as a separate compression vector.  
We adopt a \emph{curriculum learning} strategy~\citep{bengio2009curriculum,wang2021survey} to improve training stability (Appendix~\ref{app:implementation_details}). 

\noindent \textbf{Context Reconstruction.} 
After learning to decode individual compression vectors, we extend the model to full context reconstruction: 
\begin{equation}
\mathcal{L}_{\mathrm{recons}} (c)= - \log P_{\theta} (c \mid \{\mathrm{Enc}(s_i), \forall s_i \in c\}, x_{\text {ins}}). 
\label{eq:context_reconstruction}
\end{equation}
Here, $c$ is a document composed of multiple chunks $\{s_i\}$, each encoded as a separate vector. 
Unlike traditional extractive or abstractive summarization methods~\citep{xu2024recomp} that require multiple passes, these vectors naturally serve as high-ratio, parallelizable summaries. 

\textbf{Training Corpus Selection.} Since the goal is to align the embedding spaces, the pretraining corpus is domain-agnostic and can be drawn from any natural language dataset. 
We use the Wikipedia dataset~\citep{izacard2023atlas}, which provides broad topical diversity and diverse narrative styles, and has proven effective for language model pretraining~\citep{gao2023enabling}.  
In Section~\ref{sec:decode_compress_tokens} and Tables~\ref{tab:decode_compress_token_mistral7b}/\ref{tab:context_reconstruction_mistral7b}, we demonstrate that these compression vectors are able to encode detailed information, such as exact organization names, academic terms, and numeric values. 

\vspace{-2mm}
\subsection{Instruction-tuning and Inference}
\label{sec:instruction_tuning}
\vspace{-2mm}

Simple `retrieve‐and‐read' pipelines often implies redundant evidence and overlook interdependencies between previously retrieved and newly needed information~\citep{wangsearching}. 
In long-context understanding, \emph{what} should be retrieved next hinges on \emph{what} has already been inferred from previously retrieved evidence~\citep{sarthi2024raptor,li2024graphreader}. 
To address this, \method 
leverages a 2-stage context refinement, which 
interleaves \emph{retrieval} and \emph{reasoning}: 1) a \emph{coarse} retrieval step eliminating irrelevant documents while maintaining computational efficiency; 
2) a \emph{fine-grained} reranking step that iteratively refines contexts for informativeness, relevance, and diversity. 


\noindent \textbf{Instruction-tuning.} 
Initially, \method is instruct-tuned to holistically reason over both formats--the top-$k$ passages are input as natural text, while the remaining are passed as compression vectors (Figure~\ref{fig:model}). 
For faster training, we instruct-tune the LLM generator on downstream tasks with LoRA~\citep{hu2021lora} using top-$n$ contexts retrieved via BM25~\citep{robertson2004simple}.

\noindent \textbf{Dynamic Evidence Reranking.} 
Effective RAG requires balancing \emph{relevance}--which ensures alignment with the user query--and \emph{novelty}--which introduces new information beyond existing evidence. 
To achieve this, we adopt an iterative evidence selection method (Algorithm~\ref{alg:evidence_selection}) that dynamically selects context based on its incremental value to model understanding. 

\emph{Embedding-based Novelty} ranks candidates based on their contribution to the model's discrepancy in knowledge, selecting the vector that minimizes the discrepancy between the selected set $\mathcal{V}_{\mathrm{sel}}$ with the query representation $\mathbf{v}_q$ in the embedding space: 
\begin{equation}
\mathrm{SelectEvi}(q, \mathcal{V}_{\mathrm{sel}}, \mathcal{V}) = \underset{v_i \in \mathcal{V} \setminus \mathcal{V}_{\mathrm{sel}}}{\mathrm{argmin}} \left\| \mathbf{v}_q - \mathrm{Aggregate}\left( \{ \mathrm{Enc}(v) \mid v \in \mathcal{V}_{\mathrm{sel}} \cup \{v_i\} \} \right) \right\|_2.
\label{eq:embed}
\end{equation} 
Since the user query is usually succinct, we supplement the query representation $\mathbf{v}_q$ by aggregating the embeddings of both the question and the top-$1$ retrieved context: $\mathbf{v}_q = \mathrm{Avg}(\mathrm{Enc}(q), \mathrm{Enc}(v_1))$.

\emph{Conditional Self-information (CSI).} An alternative is to select evidence based on CSI~\citep{shannon1948mathematical}, which quantifies the surprisal of new evidence given previously selected evidence: 
\begin{align}
\mathrm{SelectEvi}(q, \mathcal{V}_{\mathrm{sel}}, \mathcal{V}) &= \underset{v_i \in V \backslash \mathcal{V}_{\mathrm{sel}}}{\mathrm{argmax}} 
\label{eq:csi}
I(v_i|\mathcal{V}_{\mathrm{sel}}) \\
I(v_i|\mathcal{V}_{\mathrm{sel}}) &= \frac{1}{|v_i|} \sum_{j=1}^{|v_i|} -\log P(w_i^j \mid v_i \in \mathcal{V}_{\mathrm{sel}}, w_i^1, \ldots, w_i^{j-1})
\label{eq:csi_calculation}
\end{align}
where $I(v_i|\mathcal{V}_{\mathrm{sel}}) = -\log P(v_i|\mathcal{V}_{\mathrm{sel}})$ is the conditional self-information of context $v_j$ given selected contexts $\mathcal{V}_{\mathrm{sel}}$, estimated using a smaller proxy language model. 
Higher CSI introduces novel information, while lower CSI suggests redundancy with previously selected content. Filtering low-CSI candidates reduces repetition and enhances context \emph{diversity} with minimal impact on overall informativeness.

\input{algo/alg1}


%% file: algo/alg1.tex
\begin{algorithm}[ht]
\caption{Query Expansion and Novelty-Based Evidence Selection.}
\label{alg:evidence_selection}
\textbf{Input:} Corpus $\mathcal{C} = \{v_i\}_{i=1}^{|\mathcal{C}|}$, query $q$, number of top contexts $n, k$ \\
\textbf{Output:} Ranked evidence set $\mathcal{V}_{\mathrm{sel}}$
\begin{algorithmic}[1]
    \State $\mathcal{V} = \mathrm{Retriever}(q, \mathcal{C})$ \Comment{Retrieve top $n$ contexts.}
    \State $\mathbf{v}_q = \mathrm{Avg}(\mathrm{Enc}(q), \mathrm{Enc}(v_1))$ \Comment{Initialize query embedding with top-1 retrieval $v_1$.}
    \State $\mathcal{V}_{\text{sel}} \leftarrow \{v_1\}$ \Comment{Initialize the set of selected contexts.}
    \For{$j = 2$ to $k$}
        \State 
        $\hat{\mathbf{v}} = \mathrm{Aggregate}(\mathrm{Enc}(v), v \in \mathcal{V}_{\text{sel}})$
        \Comment{Aggregate embeddings of $\mathcal{V}_{\text{sel}}$.}
        \State 
        $v_i^{\ast} = \mathrm{SelectEvi}(q, \mathcal{V}_{\mathrm{sel}}, \mathcal{V})$
        \Comment{Evaluate and select context via Eq.~\ref{eq:embed} or \ref{eq:csi}.}
        \State $\mathcal{V}_{\mathrm{sel}} \leftarrow \mathcal{V_\mathrm{sel}} \cup \{v_i^*\}$ \Comment{Update the selected context set.}
    \EndFor
    \State \Return $\mathcal{V}_{\mathrm{sel}}$
\end{algorithmic}
\end{algorithm}

%% file: src/eval.tex
\vspace{-3mm}
\section{Evaluation}
\vspace{-2mm}
\subsection{Experimental Setup}
\noindent \textbf{Baselines.} We compare our methods with $8$ baselines spanning $3$ categories:
1) \emph{Standard RAG}~\citep{lewis2020retrieval}, which directly feed retrieved documents to the input prompt; 
2) \emph{Compression-based methods}, which condense input passages before feeding them into the LLM, including LLMLingua~\citep{jiang2023llmlingua}, LongLLMLingua~\citep{jiang2023longllmlingua}, ICAE~\citep{gecontext}, and xRAG~\citep{cheng2024xrag};
3) \emph{Summarization-based methods}, which generate intermediate summaries over retrieved documents to support more focused reasoning, including Raptor~\citep{sarthi2024raptor},  GraphRAG~\citep{edge2024local}, and InstructRAG~\citep{wei2024instructrag}. 
For summarization-based approaches such as Raptor and GraphRAG, which rely on community-level summarization and long-context reasoning, we adopt the more powerful GPT-4o~\citep{gpt4o} as the base model, following prior work~\citep{luo2025semi, li2025grappi}, as open-source models struggle with reasoning over long complex inputs. 

\noindent \textbf{Generalizability Experiments.} To demonstrate the modularity and robustness of our approach, we evaluate its generalizability across different retrieval, embedding, and generation components. For the \emph{retrieval} module, we experiment with both sparse and dense retrievers, including BM25~\citep{robertson2004simple}, \texttt{bge-reranker-v2-m3}~\citep{li2023making} and \texttt{SFR-Embedding}~\citep{SFRAIResearch2024}.

\noindent \textbf{Datasets.} We evaluate our approach across diverse datasets spanning different domains, input length, and task types:  
1) \emph{Short-context question answering}, including SQuAD-v2.0~\citep{rajpurkar2018know}
2) \emph{Long-context question answering}, which requires responses based on a single long document, including NarrativeQA~\citep{kovcisky2018narrativeqa}, QASPER~\citep{dasigi2021dataset}, QuALITY~\citep{pang2022quality}, and MultifieldQA-en~\citep{bai2024longbench}; 
3) \emph{Multi-hop reasoning}, which requires multi-hop inference across documents, including HotpotQA~\citep{yang2018hotpotqa}, TriviaQA~\citep{joshi2017triviaqa},  2WikiMultihopQA~\citep{ho2020constructing};  
4) \emph{Summarization}, including QMSum~\citep{zhong2021qmsum}.  
We use SQuAD-v2, NarrativeQA, QASPER, QuALITY, HotpotQA, and TriviaQA for both training and evaluation. MultifieldQA-en, 2WikiMultihopQA, and QMSum are held out for out-of-domain evaluation only.
Detailed dataset descriptions and statistics are in Appendix~\ref{app:dataset}.

\input{tab/tab-overall}

\noindent \textbf{Metrics.} 
We adopt standard evaluation protocols consistent with prior work~\citep{asai2023self,cheng2024xrag,sarthi2024raptor,edge2024local}. 
For holistic evaluation, we report both traditional \emph{lexical metrics}--including ROUGE (R-L)~\citep{lin2004rouge}, F1 match scores--and \emph{LLM-based metrics}~\citep{es2024ragas}, including response relevance, answer correctness, semantic similarity, and faithfulness. Full metric definitions and implementation details are in  Appendix~\ref{app:metrics}.


\input{tab/tab-overall_compress_llm_metrics}

\vspace{-3mm}
\subsection{Overall Performance}
\label{sec:overall}
\vspace{-2mm}
\input{tab/tab-overall_compress_methods}

\noindent \textbf{Results under Context Constraints.} 
Tables~\ref{tab:overall_compress_llm_metrics} and~\ref{tab:overall_compress} compare \method and strong compression-based methods under strict context length constraints ($512$ and $1024$ tokens). 
\method consistently outperforms baselines on both lexical (F1, ROUGE-L) and LLM-based evaluation metrics. 
Under 512 tokens, \method improves F1 by 19.4\% and ROUGE-L by 20.8\% on average. We observe that the gains are  particularly significant on knowledge-intensive tasks like TriviaQA (+24.5\%) and HotpotQA (+29.0\%), which require facts and reasoning. Improvements on narrative-style tasks (e.g. NarrativeQA) are more modest, particularly under 1024 tokens (+6.6\% F1 and 6.8\% ROUGE-L), likely because chunking and compression can change the narrative flow and obscure subtle discourse-level cues. 
Unlike factoid questions, 
narrative questions demand holistic coherence that is harder to retain under chunking and summarization~\citep{gecontext}. 

\noindent \textbf{Impact of Context Budgets.} 
Increasing the context budget from 512 to 1024 tokens generally improves performance. 
Baselines that produce natural language compression (e.g., LongLLMLingua) see substantial gains--up to $+10.6$ F1 on NarrativeQA--as the additional budget reduces the need to truncate or overly compress passages, allowing inputs to better reflect their original structure. 
\method retains a clear performance lead, outperforming the strongest baseline by 6-12 F1 on knowledge-intensive tasks such as TriviaQA and HotpotQA. 
\input{tab/tab-SQuADv2}
As \method has already captured key content efficiently under a lower context budget using its hybrid compression strategy, it exhibits relatively modest gains on certain datasets (e.g., +4.1 F1 on QASPER). 

\noindent \textbf{Balancing Compression Efficiency and Answer Faithfulness.} 
A central challenge in RAG is balancing compression efficiency with faithfulness. Aggressive approaches like xRAG, which compress entire evidence sets into a single dense vector, optimize for efficiency but often at the cost of factuality and hallucination.  
As shown in Table~\ref{tab:overall_performance}, baselines like xRAG especially struggle on knowledge-intensive tasks, achieving only $43.4$ F1 and $35.5$ ROUGE-L on TriviaQA, in contrast to \method's $85.1$ F1 and $83.9$ ROUGE-L. Qualitative analysis in Table~\ref{tab:answers_compress_methods} reveals that baselines can hallucinate content, generating answers with fabricated entities or tasks (`sentiment analysis' and `machine translation') ungrounded in the original documents. 
Methods that over-compress inputs (e.g. ICAE) risk discarding critical content. As a result, the model tends to become overly conservative-frequently concluding that the answer is not present. 
These failures underscore the drawbacks of one-shot compression when multiple facts must be retained. 
In contrast, \method can accurately recovers fine-grained content, such as specific task names (e.g. NLI, document and intent classification) prompted in the question) with high fidelity, even under tight context budgets. 
Thus, \method's hybrid approach preserves salient content, simplifying key information while mitigating factual distortion under tight context budgets.



\noindent \textbf{Comparison with Summarization-based methods}
\method consistently outperforms standard RAG and state-of-the-art summarization-based baselines, 
including Raptor and GraphRAG, despite their use of stronger base models like GPT-4o~\citep{gpt4o} for question-answering and summarization. 
On HotpotQA, which requires multi-hop reasoning, \method achieves +15\% F1 and +14.6\% ROUGE-L. 
These results highlight the effectiveness of our compression approach in helping the model accommodate and reason over multiple discrete evidence pieces within constrained context windows. 


\noindent \textbf{Performance on Short-context QA.} 
SQuAD-v2 presents minimal challenges in context length, as each query is paired with a single passage that fits within the model's input window in most cases. Accordingly, the performance gap across models narrows. 
\method achieves the highest results (76.55 F1, 69.22 ROUGE-L; Table~\ref{tab:SQuADv2}), outperforming the strongest baseline by a modest margin (+3.98 F1, +2.19 ROUGE-L). 
In contrast, aggressively compressed systems such as xRAG and ICAE perform significantly worse ($\le 60.19$ F1), likely due to summaries that obscures critical details--such as entity names, numeric values, and specific events--reducing accuracy even when full text fits into the model.

\vspace{-1mm}
\subsection{Generalization across LLM Architectures \& Sizes.} 
\label{app:generalization_llms}
\begin{wrapfigure}{r}{0.45\textwidth}
\centering
\includegraphics[width=0.98\linewidth]{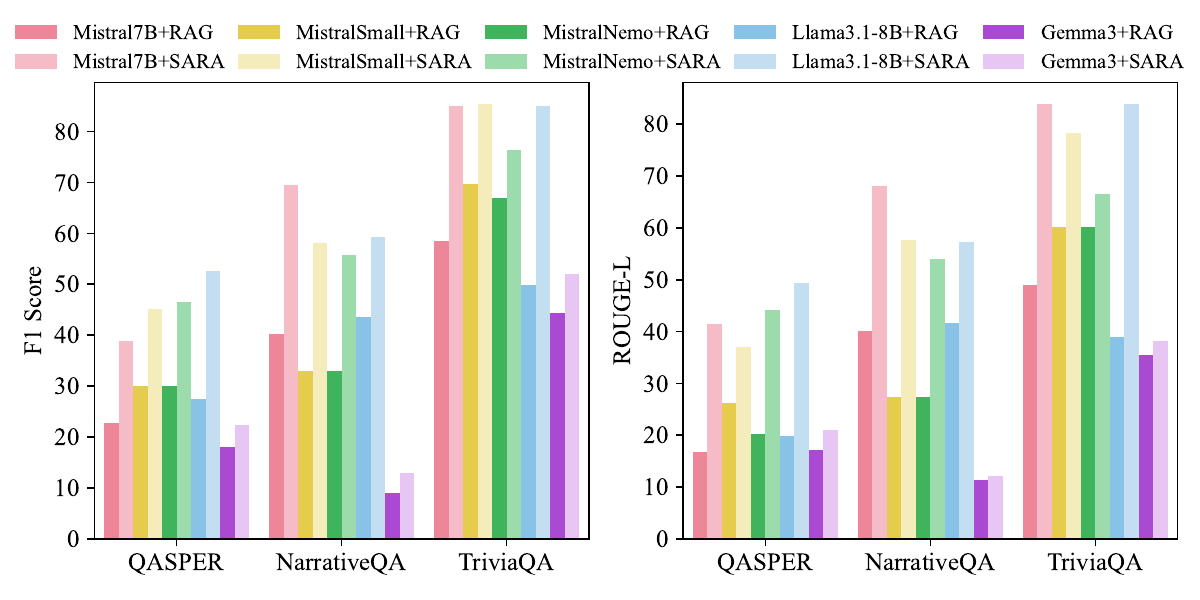}
\vspace{-3mm}
\caption{Generalizability across models. We report lexical metrics (F1 score and ROUGE-L) on QASPER~\citep{dasigi2021dataset} before and after applying \method.}
\label{fig:generalizability_f1_rougel}
\vspace{-3mm}
\end{wrapfigure}

\begin{figure}
\centering
\includegraphics[width=0.95\linewidth]{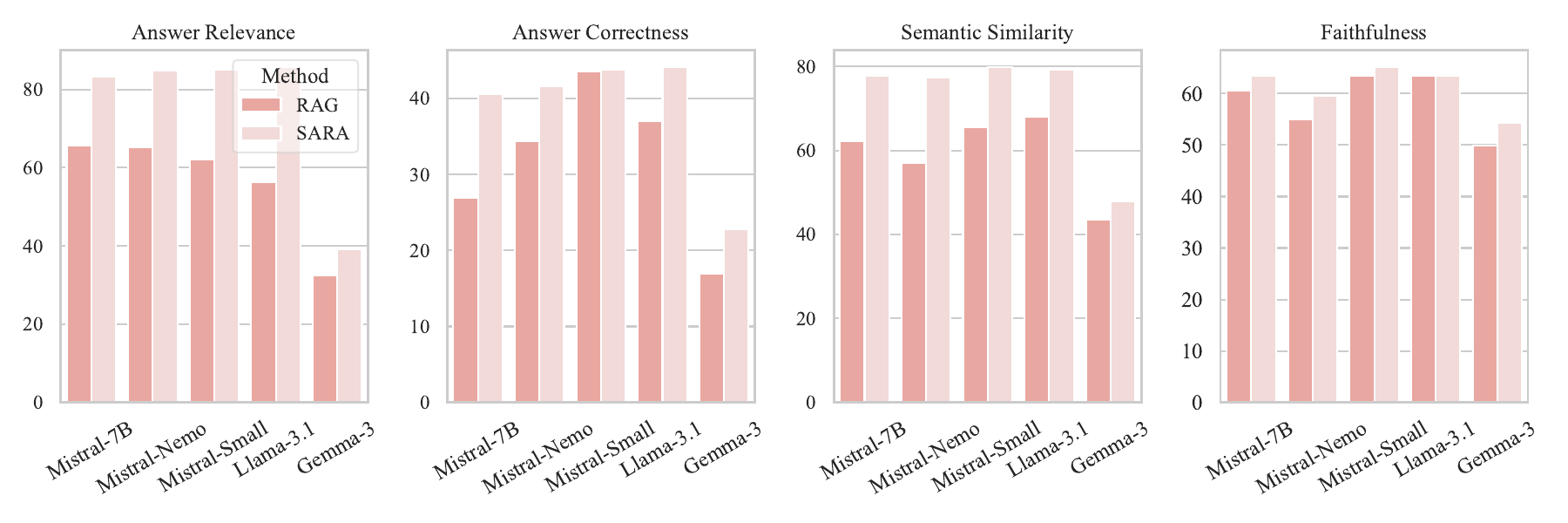}
\vspace{-3mm}
\caption{Performance of RAG and \method across different LLMs in terms of LLM-based metrics on QASPER~\citep{dasigi2021dataset}. 
}
\label{fig:generalization_base_model_llm_metrics}
\vspace{-4mm}
\end{figure}

\vspace{-1mm}
Beyond Mistral-7B, we evaluate \method on 4 additional models from 3 families--Mistral, Llama, and Gemma--spanning various sizes and architectures: MistralNemo-12B, MistralSmall-24B, Llama3.1-8B, and Gemma3-4B. 
As shown in Figures~\ref{fig:generalization_base_model_llm_metrics} and \ref{fig:generalizability_f1_rougel},  
\method consistently outperforms the baseline, with up to +40 in Answer Relevance, +14 in Answer Correctness, and +21 in Semantic Similarity. 
Improvements are particularly pronounced on smaller models. 
On Mistral-7B, \method boosts answer relevance by 17.71, answer correctness by 13.72, and semantic similarity by 15.53. 
These results highlight the method’s ability to optimize context usage under tighter context budgets, making it especially effective for smaller models. 
In some cases, \method enables a 7B model to match or surpass much larger ones (e.g., MistralSmall-24B), highlighting that reasoning over mixed-format contexts can close the performance gap without increasing model sizes. 

In general, performance gains are more significant when the compressor and LLM share the same architecture (e.g. Mistral). 
Among the Mistral family, we observe an average boost in Answer Relevance of 20.12 and Answer Correctness of 7.07. 
MistralNemo and MistralSmall achieve improvements in response relevance of +19.65 and +23.01, and semantic similarity of +20.44 and +14.38, respectively. This suggests that architectural alignment between the compressors and LLMs enhances semantic compatibility between compressed inputs and answer generation. 
In contrast, Gemma-3 shows modest gains (e.g. +6.83 in answer relevance and +5.82 in answer correctness), likely due to its architectural mismatch. 

Note that \method does not aim to directly enhance the QA model's intrinsic generation capability. Instead, its strength lies in refining and reorganizing retrieved contexts to support finer-grained reasoning. 
Since both \method and RAG leverage the same initial retriever, they operate over comparable evidence. 
As a result, faithfulness--the factual consistency with the retrieved context--shows modest improvements.

\input{tab/tab-decode_compress_token_mistral7b}

\subsection{Generalization Across Retrievers}
\label{app:generalization_retrievers}
We evaluate \method with dense retrievers like \texttt{multi-qa-mpnet-base-cos-v1}~\citep{song2020mpnet} and SFR~\citep{SFRAIResearch2024} in addition to BM25~\citep{robertson2004simple}. 
As shown in Table~\ref{tab:vary_retrievers},  \method performs consistently across retrievers, confirming its model-agnostic design. 
Dense retrievers, especially SFR, yield stronger results--achieving +19 F1 over BM25 on QASPER--highlighting the value of semantically richer base retrievers for complex, multi-hop QA. Overall, \method remains robust to retriever choice while benefiting from higher-quality evidence. 

\vspace{-3mm}
\subsection{Ablation Studies}
\vspace{-3mm}
To quantify the contribution of each major component--compression, reconstruction, and reranking--we evaluate $3$ variants of \method. 
\textbf{\method-C} removes the \textbf{C}ompression vectors and only process contexts in natural language formats. 
\textbf{\method-P} removes the context reconstruction objective during training (Section~\ref{sec:compression_learning}). 
\textbf{\method-R} skips the adaptive reranking stage, relying solely on initial BM25 retrieval (Section~\ref{sec:instruction_tuning}).

\begin{wrapfigure}{r}{0.4\textwidth}
\centering
\vspace{-5mm}
\includegraphics[width=0.97\linewidth]{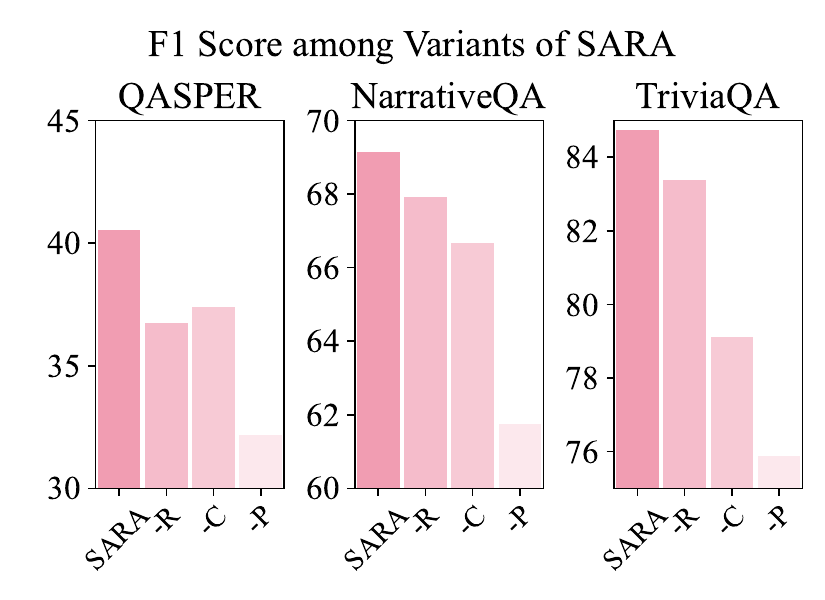}
\caption{Performance of \method's variants.}
\label{fig:ablation_f1}
\vspace{-6mm}
\end{wrapfigure}

\noindent \textbf{Context Reconstruction is Critical.} 
Removing the reconstruction objective  (\method-P) results in the most substantial performance drop (Figure~\ref{fig:ablation_f1})--7-9 F1 across all datasets. 
This confirms that learning to reconstruct full contexts from compressed vectors is essential for preserving semantic and leveraging these vectors for accurate answer generation.

\noindent \textbf{Compression Enhances Robustness.} 
Disabling compression (\method-C) also leads to consistent performance declines, especially on TriviaQA (-5.6 F1) where the long-form contexts are potentially noisy or irrelevant. 
Compression helps filter salient content and suppress redundancy, enhancing answer correctness. 

\noindent \textbf{Reranking offers Measurable Gains.} 
Removing reranking (\method-R) yields modest but consistent drops, confirming that compression-aware reranking improves evidence selection beyond lexical similarity--especially when initial retrieval are suboptimal--at minimal computational cost.


\begin{figure}[htbp] 
\centering
\includegraphics[width=0.19\linewidth]{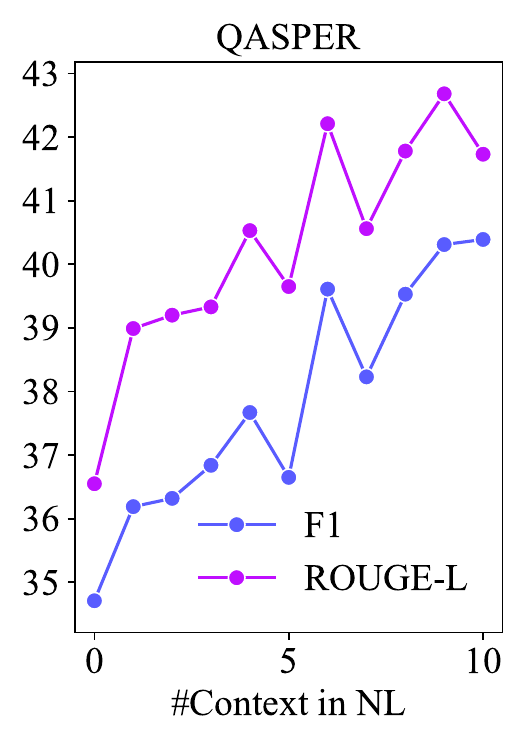}
\includegraphics[width=0.19\linewidth]{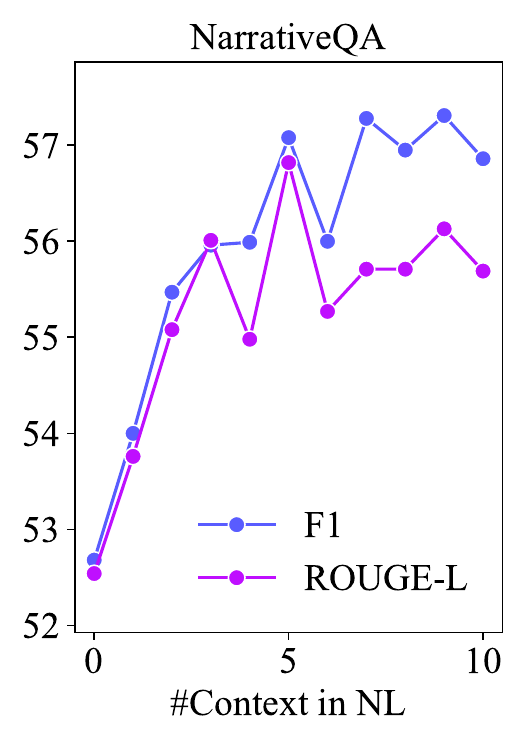}
\includegraphics[width=0.19\linewidth]{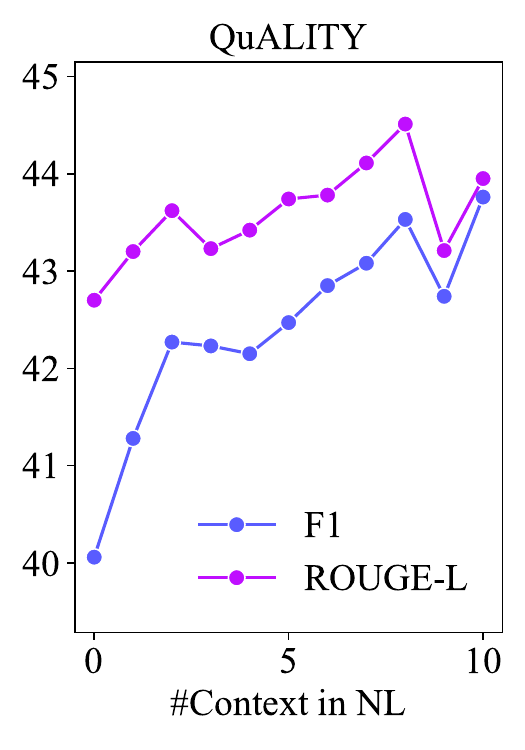}
\includegraphics[width=0.19\linewidth]{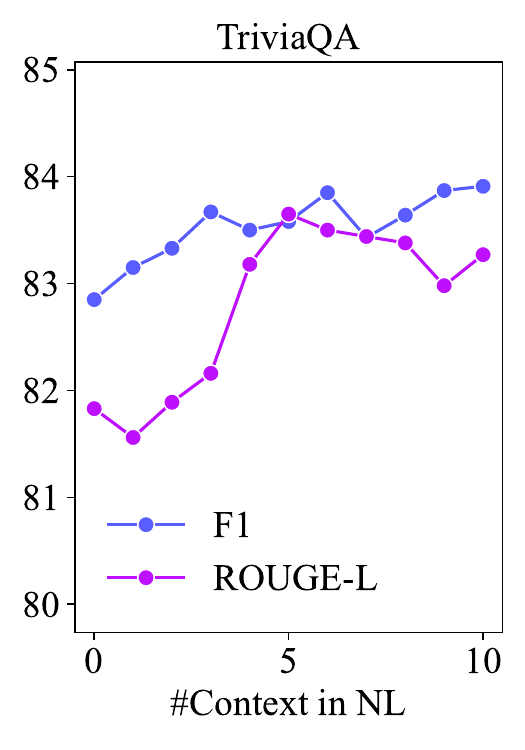}
\includegraphics[width=0.19\linewidth]{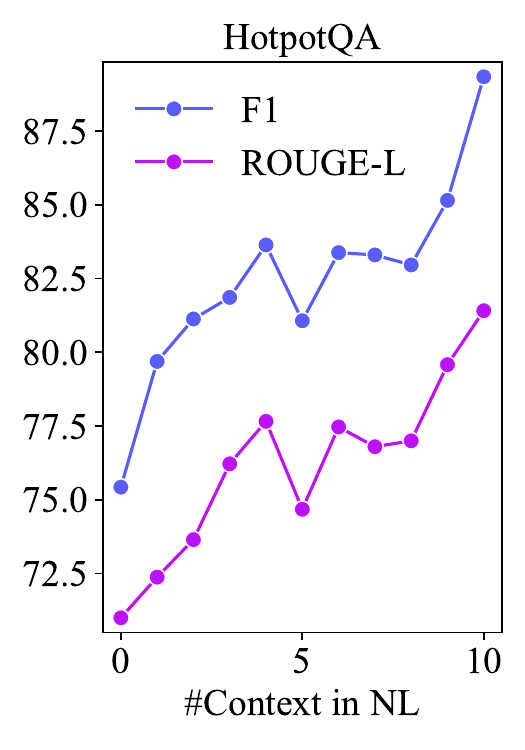}
\vspace{-1mm}
\caption{Sensitivity analysis with total contexts fixed at $N=10$, varying the number of natural language contexts $k$. Performance improves as $k$ increases, peaking around $k=7$-$8$, and slightly declines beyond $8$. \method achieves strong performance by optimally balancing natural language and compressed contexts, effectively minimizing token overhead without sacrificing accuracy.}
\label{fig:sensitivity}
\vspace{-5mm}
\end{figure}

\subsection{Sensitivity Analysis}
We evaluate \method's ability to leverage compressed context by fixing the total number of retrieved contexts ($N=10$) and varying $k$, the number of top-ranked passages retained in natural language. 
As shown in Figure~\ref{fig:sensitivity}, performance remains strong even with minimal natural language input (e.g., $k=1$, F1$=38.54$, ROUGE-L$=39.89$), indicating that compression vectors retain essential information.
Performance improves with larger $k$ but plateaus around $k=8$ (F1$=41.6$, ROUGE-L$=43.12$), and slightly drops at $k=9$, suggesting diminishing returns or noise from excessive natural language content. 
These results highlight the effectiveness of our hybrid strategy in balancing context utility, informativeness, and efficiency. 

To further illustrate such effects, Table~\ref{tab:sensitivity_answers} shows how increasing $k$ within a specific range improves factual specificity. With only compressed context (0/10), the model is able to identify a single entity name (CoNLL-2003), whereas increasing $k=5$ enables the model to produce answers with high fidelity. Our hybrid approach allows for such precision without overwhelming the context budget.

%% file: tab/tab-overall.tex
\begin{table*}[!ht]
\centering
\setlength{\tabcolsep}{3.5pt}
\begin{tabular}{lcccccccccc}
\toprule
Dataset & \multicolumn{2}{c}{QASPER} & \multicolumn{2}{c}{NarrativeQA} & \multicolumn{2}{c}{TriviaQA}& \multicolumn{2}{c}{QuALITY} & \multicolumn{2}{c}{HotpotQA} \\ 
Metrics & F1 & R-L & F1 & R-L & F1 & R-L & F1 & R-L & F1 & R-L \\ 
\midrule
RAG & 22.73 & 16.71 & 40.23 & 40.16 & 58.43 & 49.07 & 31.79 & 31.63 & 48.56 & 40.06 \\ 
Raptor & 31.77 & 25.26 & 56.60 & 56.91 & 70.51 & 65.46 & 34.27 & 34.49 & 68.26 & 63.14 \\ 
GraphRAG & 37.05 & 36.66 & 64.93 & 63.55 & 77.52 & 72.35 & 37.21 & 38.15 & 73.23 & 68.21 \\ 
xRAG & 32.36 & 33.72 & 33.43 & 32.15 & 43.36 & 35.52 & 32.65 & 33.84 & 60.19 & 49.56 \\
InstructRAG & 32.83 & 33.92 & 41.79 & 39.85 & 76.47 & 72.19 & 37.98 & 38.30 & 66.77 & 60.18 \\ 
\midrule
\method-CSI & 38.83 & 41.52 & \textbf{69.46} & \textbf{68.02} & \textbf{85.08} & 83.85 & \textbf{42.78} & 44.18 & \textbf{84.21} & \textbf{78.16} \\ 
\method-EMB & \textbf{40.55} & \textbf{41.71} & 69.15 & 66.55 & 84.74 & \textbf{84.17} & 42.59 & \textbf{44.31} & 83.77 & 76.37 \\
\midrule
\emph{Impr. \%} & 9.4\% & 13.8\% & 7.0\% & 7.0\% & 9.8\% & 16.3\% & 12.6\% & 15.7\% & 15.0\% & 14.6\% \\
\bottomrule
\end{tabular}
\caption{Performance of \method, vanilla RAG, and state-of-the-art summarization-based methods. 
}
\label{tab:overall_performance}
\vspace{-2mm}
\end{table*}

%% file: tab/tab-overall_compress_llm_metrics.tex
\begin{table*}[ht]
\centering
\vspace{-1mm}
\begin{tabular}{lcccc|cccc}
\toprule
\multirow{2}{*}{Model} & \multicolumn{4}{c|}{QASPER} & \multicolumn{4}{c}{QuALITY} \\
\cmidrule(lr){3-4} \cmidrule(lr){7-8}
 & Rele. & Correct. & Sim. & Faith. & Rele. & Correct. & Sim. & Faith. \\
\midrule
ICAE & 75.45 & 24.03 & 59.48 & 21.72 & 63.33 & 22.18 & 59.84 & 31.05 \\
LLMLingua & 79.83 & 23.97 & 61.08 & 25.31 & 85.58 & 36.06 & 79.61 & 41.19 \\
LongLLMLingua & 82.77 & 22.86 & 62.17 & 29.77 & 86.87 & 38.90 & 83.09 & 40.86 \\
SARA & 85.35 & 25.74 & 63.99 & 31.95 & 89.23 & 49.71 & 83.51 & 43.57 \\
\bottomrule
\end{tabular}
\caption{Evaluation results across QASPER and QuALITY with a context length budget of $512$ tokens. We report Response Relevance (Rele.), Answer Correctness (Correct.), Semantic Similarity (Sim.), and Faithfulness (Faith.) in percentages. Results on other datasets are in Appendix Table~\ref{tab:overall_compress_llm_metrics_2}. 
}
\label{tab:overall_compress_llm_metrics}
\vspace{-3mm}
\end{table*}

%% file: tab/tab-overall_compress_methods.tex
\begin{table*}[ht]
\centering
\setlength{\tabcolsep}{3pt}
\begin{tabular}{l*{5}{cc}}
\toprule
\multirow{2}{*}{512 tokens} & \multicolumn{2}{c}{QASPER} & \multicolumn{2}{c}{NarrativeQA} &
\multicolumn{2}{c}{TriviaQA} & \multicolumn{2}{c}{QuALITY} &
\multicolumn{2}{c}{HotpotQA} \\
\cmidrule(lr){2-3}\cmidrule(lr){4-5}\cmidrule(lr){6-7}\cmidrule(lr){8-9}\cmidrule(lr){10-11}
  & F1 & R-L & F1 & R-L & F1 & R-L & F1 & R-L & F1 & R-L \\ 
\midrule
ICAE              & 26.64 & 23.53 & 37.58 & 38.08 & 53.47 & 49.16 & 26.79 & 28.20 & 53.18 & 44.05 \\
LLMLingua         & 31.29 & 32.38 & 50.26 & 48.58 & 63.22 & 58.95 & 30.53 & 31.48 & 57.36 & 49.30 \\
LongLLMLingua     & 29.49 & 28.31 & 41.90 & 39.27 & 66.28 & 62.76 & 36.13 & 38.03 & 64.34 & 60.32 \\
\method & \textbf{36.23} & \textbf{39.17} & \textbf{55.64} & \textbf{54.90} &
\textbf{82.50} & \textbf{81.74} & \textbf{42.27} & \textbf{43.62} & \textbf{83.03} & \textbf{75.56} \\
\midrule
\emph{Impr. \%} & 15.8 & 21.0 & 10.7 & 13.0 & 24.5 & 30.2 & 17.0 & 14.7 & 29.0 & 25.3 \\
\midrule
\multirow{2}{*}{1024 tokens} & \multicolumn{2}{c}{QASPER} & \multicolumn{2}{c}{NarrativeQA} &
\multicolumn{2}{c}{TriviaQA} & \multicolumn{2}{c}{QuALITY} &
\multicolumn{2}{c}{HotpotQA} \\
\cmidrule(lr){2-3}\cmidrule(lr){4-5}\cmidrule(lr){6-7}\cmidrule(lr){8-9}\cmidrule(lr){10-11}
  & F1 & R-L & F1 & R-L & F1 & R-L & F1 & R-L & F1 & R-L \\ 
\midrule
ICAE  & 31.82 & 33.32 & 36.70 & 38.35 & 51.07 & 49.78 & 28.15 & 29.88 & 64.51 & 55.60 \\
LLMLingua         & 33.18 & 32.19 & 50.09 & 52.46 & 71.92 & 67.01 & 33.82 & 34.90 & 62.80 & 60.71 \\
LongLLMLingua     & 34.09 & 33.47 & 52.48 & 51.17 & 72.64 & 67.47 & 36.57 & 33.18 & 69.21 & 67.88 \\
\method & \textbf{40.37} & \textbf{42.24} & \textbf{55.96} & \textbf{56.01} &
\textbf{83.67} & \textbf{82.16} & \textbf{42.40} & \textbf{44.19} & \textbf{83.77} & \textbf{76.37} \\
\midrule
\emph{Impr. \%} & 18.4 & 26.2 & 6.6 & 6.8 & 15.2 & 21.8 & 15.9 & 26.6 & 21.0 & 12.5 \\
\bottomrule
\end{tabular}
\caption{Performance of compression methods under context length constraints (512/1024 tokens) in terms of F1 scores and ROUGE-L (R-L). 
Improvements over the best models are shown with \emph{Impr.\%}.}
\label{tab:overall_compress}
\vspace{-5mm}
\end{table*}

%% file: tab/tab-SQuADv2.tex
\begin{wraptable}{r}{0.35\textwidth}
\centering
\vspace{-3mm}
\setlength{\tabcolsep}{3pt}
\begin{tabular}{lcc}
\toprule
Dataset & \multicolumn{2}{c}{SQuAD-v2} \\ \cmidrule(lr){2-3}
Metrics & F-1 & R-L \\
\midrule
RAG & 63.65 & 51.26 \\
Raptor & 70.69 & 65.28 \\
GraphRAG & 74.82 & 67.36 \\
xRAG & 60.19 & 49.56 \\
InstructRAG & 67.21	& 57.94 \\
ICAE & 50.31 & 40.82 \\
LLMLingua & 70.24 & 65.12 \\
LongLLMLingua & 72.57 & 67.03 \\
\method & 76.55 & 69.22 \\
\bottomrule
\end{tabular}
\caption{Performance comparison on the SQuAD-v2 dataset.}
\label{tab:SQuADv2}
\vspace{-3mm}
\end{wraptable}

%% file: tab/tab-decode_compress_token_mistral7b.tex
\begin{table*}[ht]
\centering
\begin{tabular}{p{0.48\textwidth}|p{0.48\textwidth}}
\toprule
\textbf{Decoded Text} & \textbf{Original Text} \\
\midrule
\black{We release the code and} the \black{data.} & \black{We release the code and data.} \\
\midrule
\black{Also, we build a persuasive dialogue system to persuade people to donate to charity.} & \black{Furthermore, we also build a persuasion dialog system to persuade people to donate to charities.} \\
\midrule
\black{Rigid templates limit} \underline{creativity} and \black{diversity, resulting in loss of user engagement.} & \black{However, rigid templates lead to limited diversity, causing the user losing engagement.} \\
\midrule
\black{The generation model is good at producing diverse responses but lacks coherence.} & On the other hand, language \black{generation models can generate diverse responses but are bad at being coherent.} \\
\midrule
\underline{Collaborative} \black{end-to-end systems have been developed to a great extent} for the goal to build a user-friendly system that \black{enables participants to work together with the system to achieve a common goal.}	& 	\black{Considerable progress has been made building end-to-end dialog systems} for collaborative tasks in which \black{users cooperate with the system to achieve a common goal.} \\
\midrule
We use a \black{hierarchical annotation scheme}. This \black{generic annotation method} can be \black{applied to different tasks.} & To handle social content, we introduce a hierarchical \underline{intent} annotation scheme, which can be generalized to different \underline{non-collaborative dialog} tasks. \\
\bottomrule
\end{tabular}
\caption{Decoded text from compression vectors using \texttt{Mistral-7B-Instruct-v0.2}~\citep{jiang2023mistral} as the base model. Information omitted from one text but present in the other is \underline{underlined}. Compared to the original, \method retains concise semantics and excels at capturing high-level concepts. In some cases, it may lose fine-grained details such as specific entities and numerical values.
}
\label{tab:decode_compress_token_mistral7b}
\vspace{-8mm}
\end{table*}

%% file: src/related.tex
\vspace{-1mm}
\section{Related Work}
\vspace{-1mm}
\subsection{Retrieval-augmented Generation (RAG)}
\vspace{-3mm}
Retrieval-augmented Generation has become a standard practice for knowledge-intensive tasks. 
Instead of treating LLMs as knowledge repositories, RAG generates answers using an external knowledge base~\citep{lewis2020retrieval,sharma2024og}. This approach helps them address model knowledge cutoffs and insufficient training coverage. 
As a common challenge for RAG models, LLMs struggle to process long, chunked retrieved contexts effectively, even with extended context windows~\citep{yurankrag}.  
Recent work such as  Raptor~\citep{sarthi2024raptor}, GraphRAG~\citep{edge2024local} and GraphReader~\citep{li2024graphreader} focus on improving the \emph{retrieval} and \emph{augmentation} stages by structuring retrieved content, enhancing RAG through semantic or graph-based organization of knowledge, leading to more relevant and compact inputs for generation.

\subsection{Context Compression}
\vspace{-2mm}
Context compression is essential for reducing inference costs and maintaining language understanding capabilities in long-context~\citep{pan2024llmlingua} or multi-turn scenarios~\citep{kimcompressed}. 
Prior work approach this in two main directions: natural-language (NL)-based compression and representation-level compression. 
\emph{NL-based compression}~\citep{zhang2024adacomp,chirkova2025provence} like \textsc{AdaComp}~\citep{zhang2024adacomp}, \textsc{CompAct}~\citep{yoon2024compact}, and \textsc{EXIT}~\citep{hwang2024exit} condense prompts or histories into concise natural language summaries, typically using extractive or abstractive summarization. These methods are generally model-agnostic and applicable across both open-source and proprietary LLMs~\citep{zhu2025generalizing}. 
Representation-based methods~\citep{chevalier2023adapting, munkhdalai2024leave,louis2025oscar,louis2025pisco}, on the other hand, treat the LLM as a white box and modify attention calculation~\citep{munkhdalai2024leave}, positional encodings~\citep{jin2024llm,zhangfound}, or embeddings~\citep{cheng2024xrag}. Methods such as xRAG~\citep{cheng2024xrag}, GIST~\citep{mu2023learning},
and ICAE~\citep{gecontext} project instructions demonstrations, or the context into the language models' space. 
While compression improves efficiency, it often introduces a performance trade-off. Our work focuses on leveraging compression to improve retrieval and generation quality in RAG settings.

%% file: src/conclusion.tex
\section{Conclusion}
\vspace{-2mm}
We present \method, a unified and efficient RAG framework that enhances both retrieval and generation through structured evidence compression and adaptive document selection without significant architectural changes to the LLM. Experiments across multiple LLM backbones, retrievers, and embedding models demonstrate that \method significantly improves answer correctness and relevance.

%% file: src/appendix.tex
\section{Experimental Details}
\label{app:experimental_details}

\begin{figure}[htbp]
\centering
\vspace{-2mm}
\includegraphics[width=0.6\linewidth]{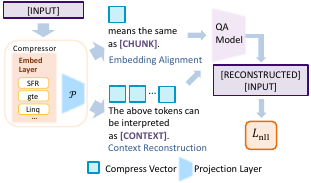}
\vspace{-2mm}
\caption{During \emph{Compression Learning}, \method learns to reconstruct text from compression vectors. 
}
\label{fig:model_pretrain}
\vspace{-2mm}
\end{figure}


\subsection{Implementation Details} 
\label{app:implementation_details}
Our implementation is based on \texttt{PyTorch}~\citep{paszke2019pytorch}, \texttt{transformers}~\citep{wolf2020transformers}, and \texttt{llama-index}~\citep{Liu_LlamaIndex_2022}. All models and data use the \texttt{bfloat16} data type. 
For LoRA setup, we adopt a rank attention dimension of $16$, scaling factor $\alpha=32$, and dropout of $0.1$. 
For chunking, we set the chunk size to $256$. The model processes at most $n=10$ chunks. Our method further selects the top $k=5$ as natural language evidence, and encode the rest as compression vectors. 
To reduce the effects of stochasticity, we fix the sampling temperature at $0$. 
Experiments were performed on a Linux server with $6$ NVIDIA A100 GPUs. 

For embedding alignment (Section~\ref{sec:compression_learning}), 
we adopt a \emph{curriculum learning} strategy, starting with shorter sentences and gradually transition into complex examples. 
Specifically, we use spaCy\footnote{\url{https://spacy.io/}} for NER and rank sentences by token count and the number of named entities in categories such as \texttt{PER}, \texttt{ORG}, \texttt{LOC}, \texttt{GPE}, \texttt{Date}, \texttt{Time}, and \texttt{Event}. The embedding models we experimented with are in Table~\ref{fig:model_pretrain}. 

\input{tab/tab-compressor_models_profile}

\subsection{Dataset Descriptions}
\label{app:dataset}

\begin{itemize}[leftmargin=1em]
    \item NarrativeQA~\citep{kovcisky2018narrativeqa}: question-answering based on books and movie transcripts.
    \item QASPER~\citep{dasigi2021dataset}: information seeking over scientific research papers with supporting evidence spans.
    \item QuALITY~\citep{pang2022quality}: reading‑comprehension benchmark with $\sim 5000$-token passages and unambiguous questions that require consolidating information from multiple text segments. 
    \item TriviaQA~\citep{joshi2017triviaqa}: trivia questions paired with web evidence (news, encyclopedia, and blogs). 
    \item HotpotQA~\citep{yang2018hotpotqa}: natural questions that require multi-hop reasoning. The questions are annotated with supporting facts. 
    \item SQuAD-v2.0~\citep{rajpurkar2018know}: questions are based on Wikipedia articles, and the answers are text segments from the corresponding reading passage. We select questions that are marked as ``answerable''
    \item QMSum~\citep{zhong2021qmsum}: query‑focused meeting summarization from dialogue transcripts.
    \item MultifieldQA-en~\citep{bai2024longbench} single-doc QA from diverse sources (arXiv, C4, Wikipedia, WuDaoCorpora, etc.) 
    \item 2WikiMultihopQA~\citep{ho2020constructing}: multi-hop QA combining structured and unstructured evidence with reasoning paths. 
\end{itemize}

\begin{figure}
\centering
\includegraphics[width=0.32\linewidth]{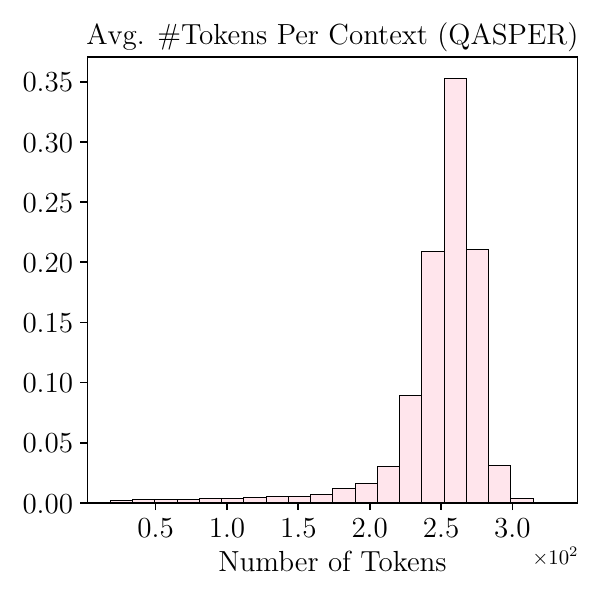}
\includegraphics[width=0.32\linewidth]{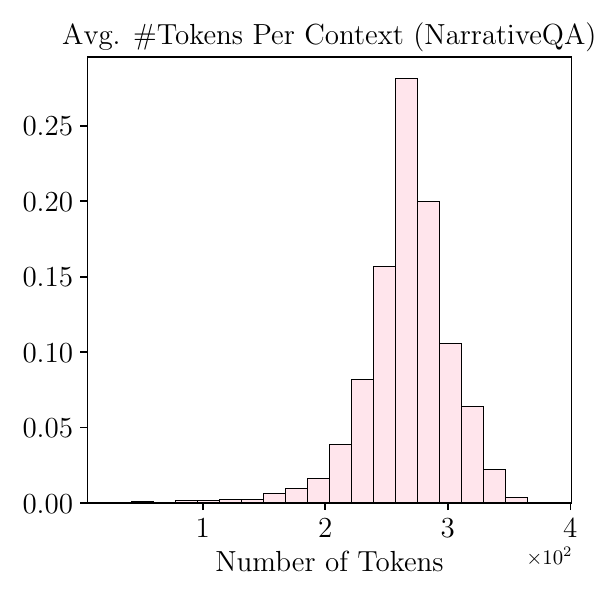}
\includegraphics[width=0.32\linewidth]{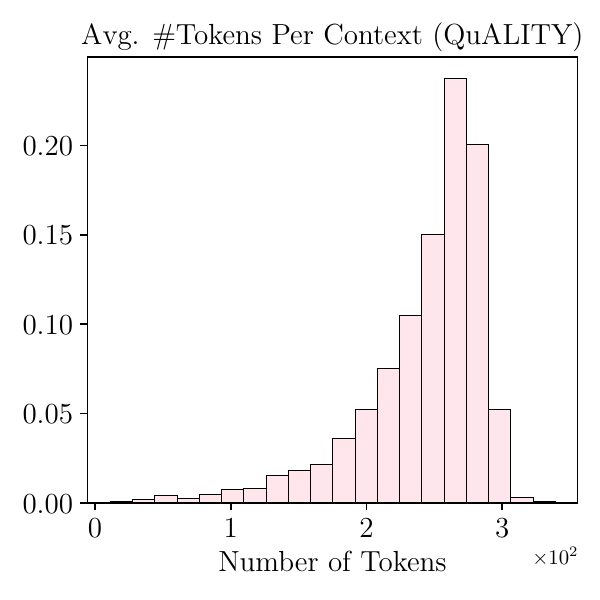}
\includegraphics[width=0.32\linewidth]{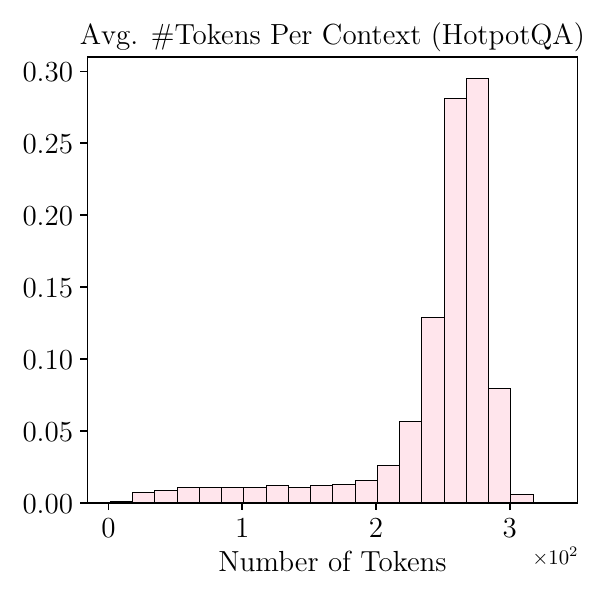}
\includegraphics[width=0.32\linewidth]{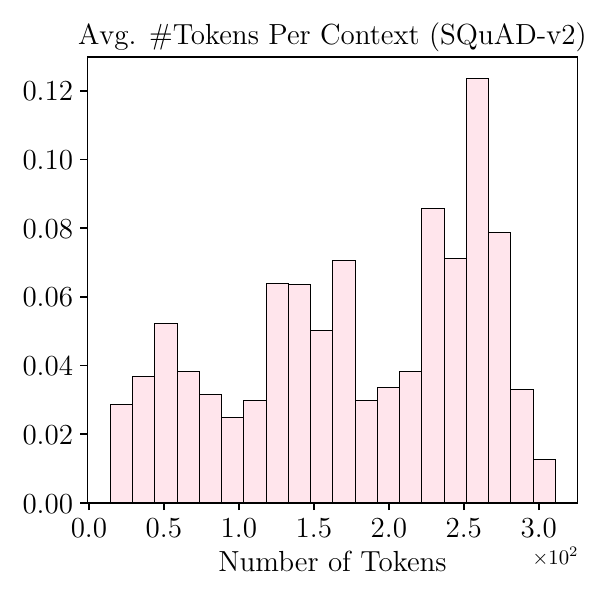}
\includegraphics[width=0.32\linewidth]{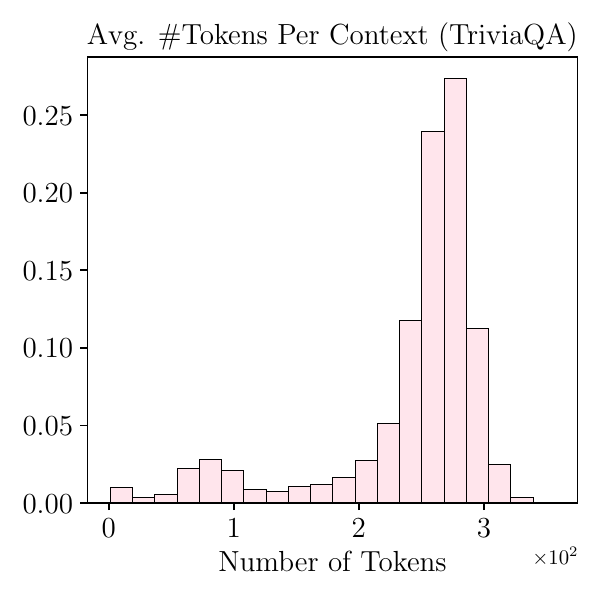}
\includegraphics[width=0.32\linewidth]{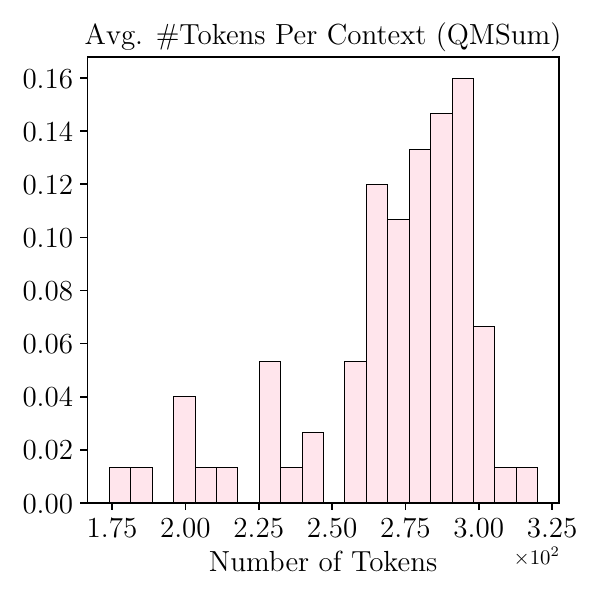}
\includegraphics[width=0.32\linewidth]{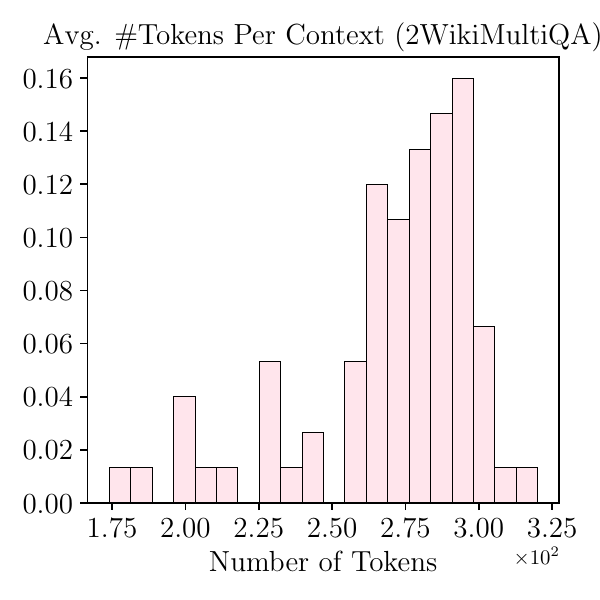}
\includegraphics[width=0.32\linewidth]{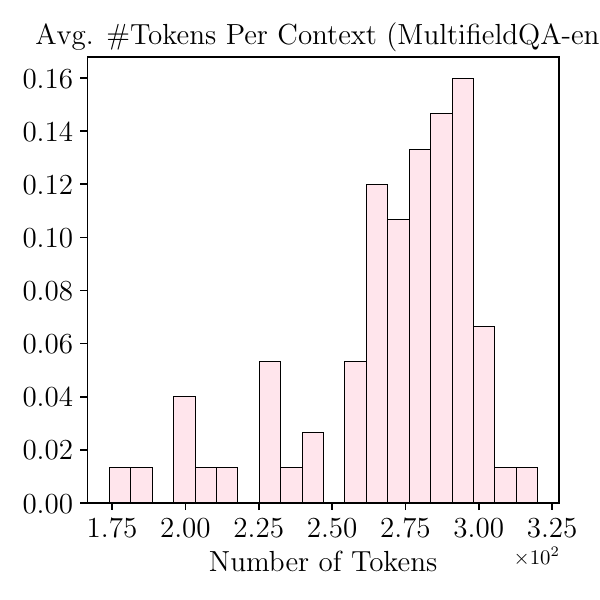}
\caption{Distribution of number of tokens per chunk in each dataset.}
\label{fig:chunk_sizes}
\end{figure}

\begin{figure}
\centering
\includegraphics[width=0.19\linewidth]{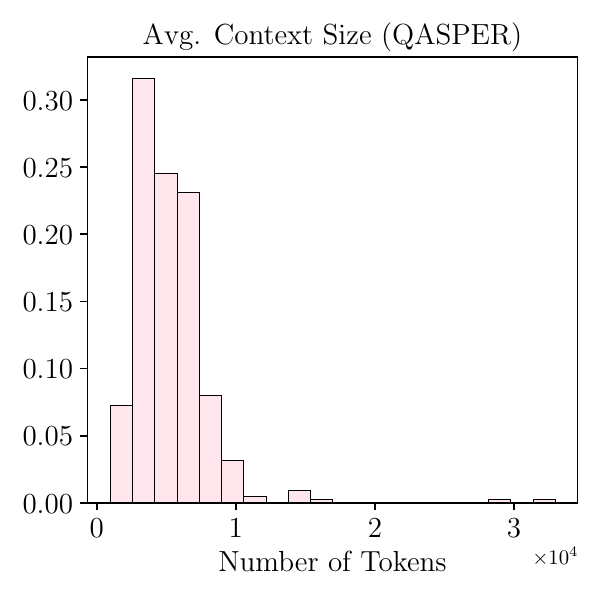}
\includegraphics[width=0.19\linewidth]{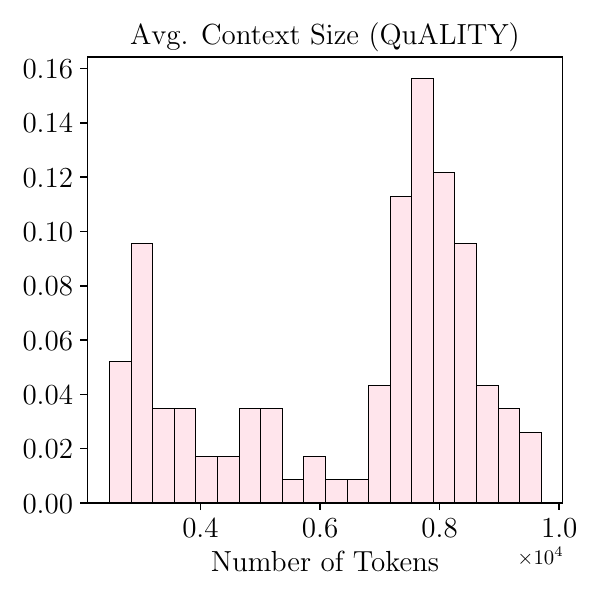}
\includegraphics[width=0.19\linewidth]{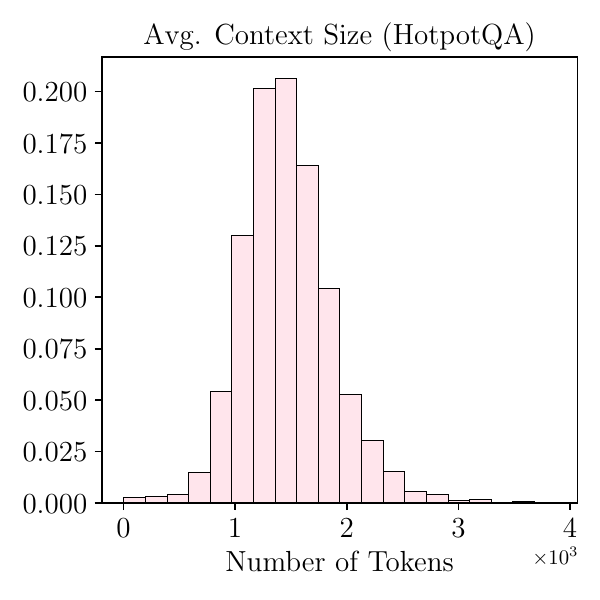}
\includegraphics[width=0.19\linewidth]{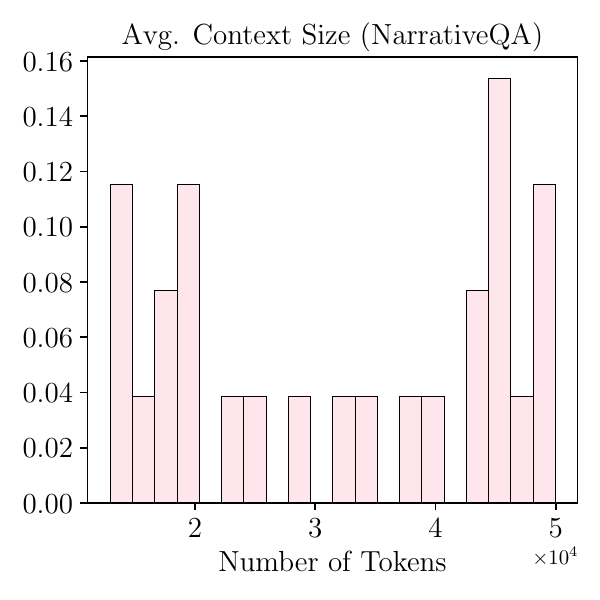}
\includegraphics[width=0.19\linewidth]{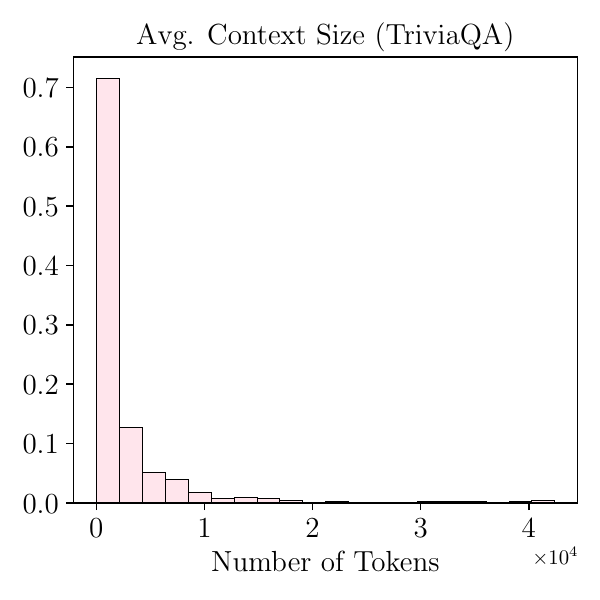}
\includegraphics[width=0.19\linewidth]{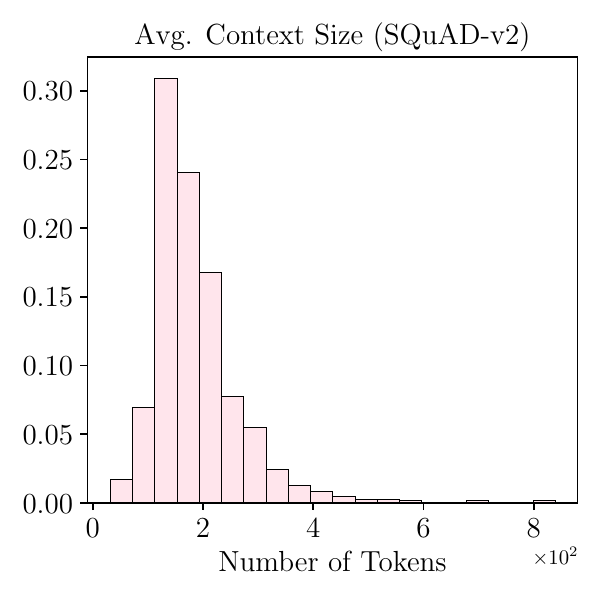}
\includegraphics[width=0.19\linewidth]{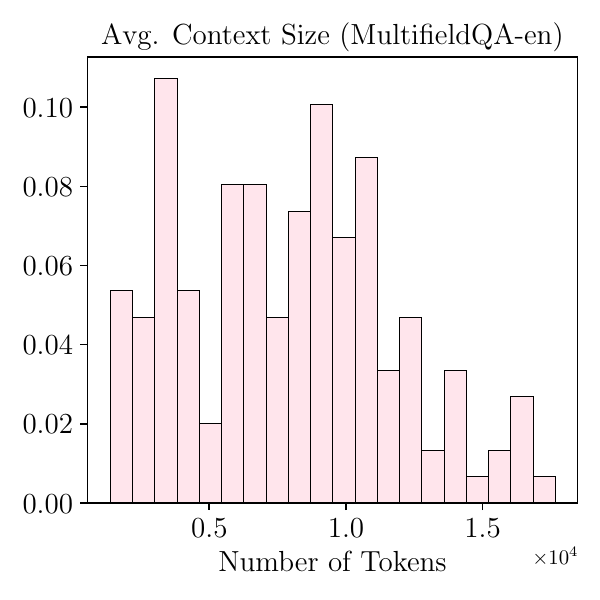}
\includegraphics[width=0.19\linewidth]{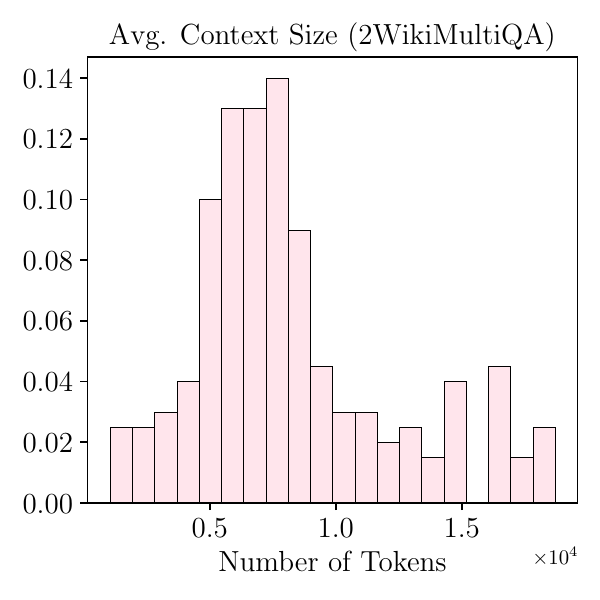}
\includegraphics[width=0.19\linewidth]{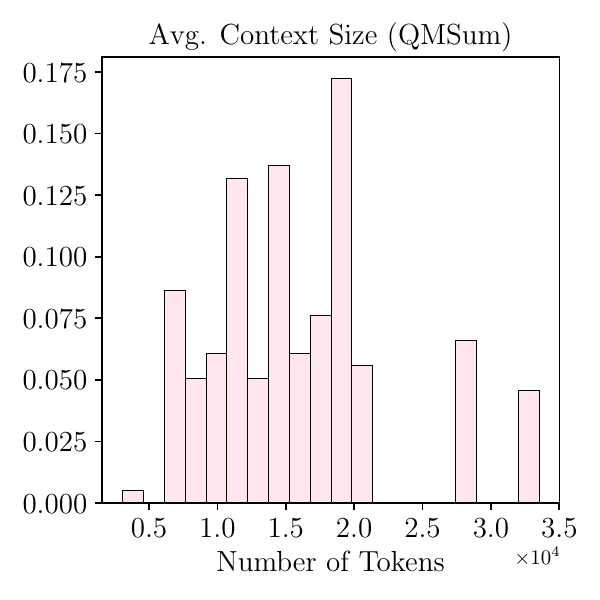}
\caption{Context Size in terms of number of tokens according to Mistral-7B's tokenizer. All datasets except SQuAD-v2 focus on long context. }
\label{fig:dataset_stats}
\end{figure}

All corpora are split into $256$‑token chunks aware of the sentence structures. The token‑count distribution is in Figure~\ref{fig:chunk_sizes}, and the overall statistics is in Figure~\ref{fig:dataset_stats}. To improve fine-tuning, we use GPT-4o~\citep{gpt4o} to convert the fine-tuning dataset into instruction-following format, following previous works~\citep{liu2023visual,xiao2024proteingpt}.

\input{tab/tab-answers_compress_methods}

\input{tab/tab-sensitivity_answers}

\input{tab/tab-prompt}

\input{tab/tab-context_reconstruction}

\input{tab/tab-overall_compress_llm_metrics_2}

\subsection{Evaluation Metrics}
\label{app:metrics}
\noindent\textbf{Automatic Evaluation.} For free-form answer generation, 
we report 
ROUGE-L (R-L)~\citep{lin2004rouge} and F1 match scores to measure lexical overlap between predicted and ground-truth answers. 

\noindent\textbf{LLM-based Evaluation.}
To complement traditional lexical scores, we adopt four LLM‑based metrics that capture orthogonal dimensions essential for reliable RAG deployment~\citep{es2024ragas,risch2021semantic}. Each metric returns a value in $[0,1]$, with higher values indicating better performance.

\begin{itemize}[leftmargin=1em]
    \item \textbf{Faithfulness} measures whether the generated answer is grounded in the retrieved context. The answer is decomposed into atomic claims with GPT‑4o. Each claim is then tested for entailment against the retrieved context. Answers fully supported by the evidence are favored, and hallucinations are penalized.
    \item \textbf{Answer Relevance} (Response Relevance) judges how directly the answer addresses the user's question. Redundant, off‑topic, or missing information lowers the score. It does not take factual accuracy into consideration.
    \item \textbf{Factual Correctness} uses claim decomposition and natural language inference to verify the model's claims against reference texts.
    \item \textbf{Semantic Similarity} uses a cross‑encoder to compute the semantic overlap between the generated answer and the ground‑truth reference. 
\end{itemize}

\section{Additional Experiments}

\begin{figure}
\centering
\includegraphics[width=0.95\linewidth]{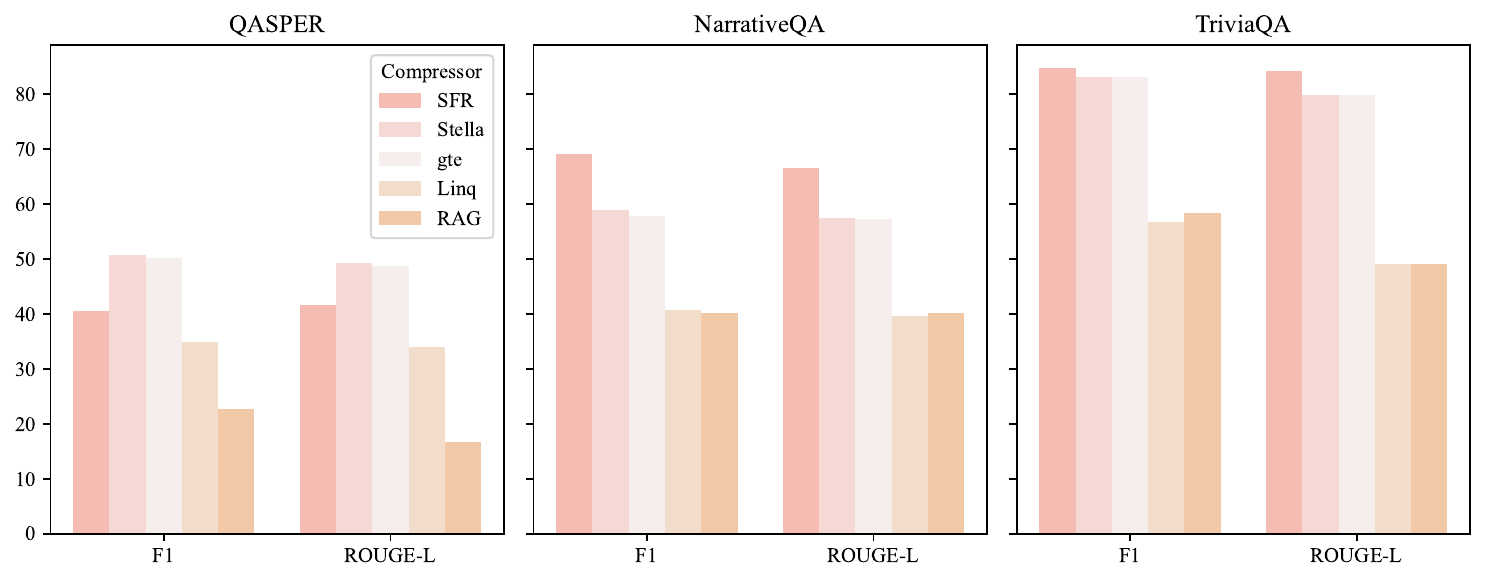}
\caption{Results on different compressors. 
}
\label{fig:generalization_compressor}
\vspace{-3mm}
\end{figure}

\input{tab/tab-vary_retrievers}

\subsection{Intrinsic Analysis of Compression Vectors}
\label{sec:decode_compress_tokens}

\subsection{Generalization on Additional Embedding Models} 
\label{app:generalization_embed}

Aside from \texttt{Salesforce/SFR-Embedding-Mistral} (SFR), we experimented with additional embeddings, including \texttt{Linq-AI-Research/Linq-Embed-Mistral} (Linq) embedding~\citep{LinqAIResearch2024}, \texttt{Alibaba-NLP/gte-Qwen2-7B-instruct}  (GTE)~\citep{li2023towards}, and \texttt{NovaSearch/stella\_en\_1.5B\_v5} (Stella). The profiles of base sentence embedding models are shown in Table~\ref{tab:compressor_models_profile}. Results are shown in Figure~\ref{fig:generalization_compressor}.

\subsection{Generalization on Unseen Datasets}
\label{app:generalization_unseen_datasets}

\input{tab/tab-ood_datasets}
We evaluate generalization by testing the fine-tuned models on three out-of-domain (OOD) datasets from LongBench~\citep{bai2024longbench}:  MultiFieldQA-en, 2WikiMultihopQA, and QMSum, which differ substantially in domain and task format from the training data (See Appendix~\ref{app:dataset} for details). 
As shown in Table~\ref{tab:ood_datasets}, \method consistently improves performance across all benchmarks. 
It boosts \textsc{Response Relevance} by wide margins--+18.5 on QMSum, +47.7 on MultifieldQA-en, and +55.0 on 2WikiMultiHopQA. 
These gains highlight the strength of combining natural language spans with compression vectors, which helps leverage more relevant evidence despite domain shifts. 
The improvements are especially pronounced on QA-style tasks, suggesting that the QA data in the fine-tuning dataset contributes to \method's performance on other QA datasets. Improved relevance also leads to cleaner answers, hallucinations and off-topic content, leading to cleaner answers. 

\begin{figure*}[htbp]
\centering
\includegraphics[width=0.4\linewidth]{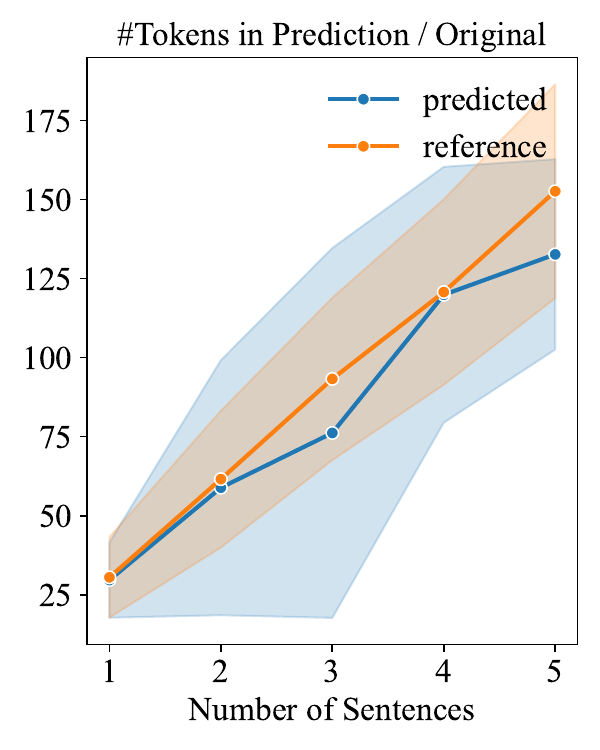}
\vspace{-1mm}
\caption{Number of words generated from compression vectors when we vary from 1 to 3 sentences. 
}
\label{fig:expressivity}
\vspace{-5mm}
\end{figure*}

In contrast, Answer Correctness rises more modestly (+0.3 to +2.2), suggesting that while retrieval quality generalizes well, reasoning over the retrieved content might be partially domain-dependent. 
For example, TriviaQA and QASPER (used in training) are based on Wikipedia and academic literature, respectively.  MultiFieldQA-en involves answering questions based on articles from multiple domains. 
In this case, in-domain adaptation or instruction tuning could help further improve this performance.

\section{Discussion}

\paragraph{Extension to New Decoders}
\method is designed to be \emph{model-agnostic}. All components--retriever, compressor, and the QA model--can be replaced with minimal effort. 
Note that the same decoder must be used across both \emph{Compression Learning} (Section~\ref{sec:compression_learning}), \emph{Instruction-tuning}, and \emph{Generation} (Section~\ref{sec:instruction_tuning}). 
This is because the model learns to \emph{interpret} compression vectors through its own decoder weights. 





\section{Expressivity of compression vectors}


Faithful representation of semantics is pivotal for our compression vectors to serve as reliable contexts. To evaluate this, we decode the compression vectors into natural language and compare the reconstructed evidence with their sources. 
Representative successes for both chunk-level and paragraph-level reconstructions are shown in Table~\ref{tab:decode_compress_token_mistral7b} and \ref{tab:context_reconstruction_mistral7b}. 
We observed that the decoded text are usually shorter and serve as higher level summarizations for the input. 
In most cases, the decoded text preserves core propositions, causal links, and sentiment. 
\method is able to recover key information, such as exact entities (e.g. `Amazon customer service') and numeric values (e.g. `220'). 
Losses are mostly fine‑grained--exact dates  (`1903' $\rightarrow$ `1900s') or numeric magnitudes (`3400 years' $\rightarrow$ over 3,000 years) may be paraphrased or omitted. 
When contexts are longer, the risk of recovery failure is higher. This necessitates reasoning over mixed evidence formats. 

Crucially, the decoder rarely invents new facts: missing detail is typically dropped rather than hallucinated. This behavior implies that the vectors encode stable, high‑level meaning while suppressing fewer specifics--a valuable feature for knowledge-intensive tasks that demand both factual precision and robust hallucination control.

%% file: tab/tab-compressor_models_profile.tex
\begin{table}[!ht]
\centering
\setlength{\tabcolsep}{3pt}
\begin{tabular}{llll}
\toprule
Model & Full Name & Base LLM & Size \\ 
\midrule
SFR~\citep{SFRAIResearch2024} & \texttt{Salesforce/SFR-Embedding-Mistral} & Mistral-7B 
& 4096 \\ 
Linq~\citep{LinqAIResearch2024} & \texttt{Linq-AI-Research/Linq-Embed-Mistral} & 
Mistral-7B 
& 4096 \\ 
GTE~\citep{li2023towards} & \texttt{Alibaba-NLP/gte-Qwen2-7B-instruct} & Qwen2-7B 
& 3584 \\ 
Stella~\citep{zhang2024jasper} & \texttt{NovaSearch/stella\_en\_1.5B\_v5} & Qwen2-1.5B 
& 8960 \\ 
\bottomrule
\end{tabular}
\caption{Embedding models used in the compressor and their embedding sizes.}
\label{tab:compressor_models_profile}
\vspace{-3mm}
\end{table}

%% file: tab/tab-answers_compress_methods.tex
\begin{table}[ht]
\centering
\begin{tabular}{p{0.17\textwidth} p{0.75\textwidth}}
\toprule
Question & How do they define rumors? \\
\midrule
xRAG & Rumors are false, unverified information that spreads quickly. \\
LLMLingua & Rumors are not explicitly defined in the context provided. \\
LongLLMLingua & They define rumors as unverified information spread on social media. \\
\methodbf & They define rumors as unverified information that spreads rapidly through a population. \\
\emph{Ground-truth} & They define rumors as the presence of rapidly spreading information that is unconfirmed by official media. \\
\midrule
Question & What are the five downstream tasks? \\
xRAG & 
1. Named Entity Recognition (NER) \newline
2. Dependency Parsing (DP) \newline
3. Sentiment Analysis (SA) \newline
4. Machine Translation (MT) \newline
5. Question Answering (QA) \\ 
LLMLingua & The five downstream tasks mentioned in the context are: machine translation, question answering, text summarization, sentiment analysis, and named entity recognition. \\
LongLLMLingua & The five downstream tasks related to grammatical error correction are text summarization, sentiment analysis, named entity recognition, question answering, and text classification. \\
\methodbf & The five downstream tasks are text / intent classification, NLI, named entity recognition, and POS tagging. \\ 
\emph{Ground-truth} & The five downstream tasks we perform using MMTE are three classification tasks: NLI (XNLI dataset), document classification (MLDoc dataset), and intent classification, as well as two sequence tagging tasks: POS tagging and NER. \\
\bottomrule
\end{tabular}
\caption{Comparison of answers generated by different compression methods.}
\label{tab:answers_compress_methods}
\end{table}

%% file: tab/tab-sensitivity_answers.tex
\begin{table}[htbp]
  \centering
  \begin{tabularx}{\textwidth}{l X}
    \toprule
    Question & Which NER dataset do they use? \\ \midrule
    \vspace{-3mm}
    \multirow{13}{*}{Evidence} & 
    \begin{itemize}[leftmargin=1em]
        \item CoNLL2003 is one of the most evaluated English NER datasets, which contains four different named entities: PERSON, LOCATION, ORGANIZATION, and MISC \ldots
        \item OntoNotes 5.0 is an English NER dataset whose corpus comes from different domains, such as telephone conversation, newswire. We exclude \ldots  
        \item \ldots OntoNotes 4.0 \ldots we use the Chinese part. We adopted the same pre-process \ldots
        \item The corpus of the Chinese NER dataset MSRA came from news domain \ldots 
        \item Weibo NER was built based on text in Chinese social media Sina Weibo, and it contained 4 kinds of entities \ldots 
        \item Resume NER was annotated by \ldots
    \end{itemize}
    \vspace{-3mm}
     \\
    \midrule
    Ground-truth & The datasets include CoNLL2003, OntoNotes 5.0, OntoNotes 4.0, the Chinese NER dataset MSRA, Weibo NER, and Resume NER. \\
    \midrule
    Predictions & \\
    0/10 & They use the \textbf{CoNLL‑2003 NER dataset}. \\
    2/8  & The NER dataset they use is \textbf{CoNLL‑2003, OntoNotes-5.0} and \gray{data based on Chinese social media.} \\
    5/5  & The NER datasets used are \textbf{CoNLL‑2003, OntoNotes-5.0, MSRA, Weibo, and Resume.} \\
    \bottomrule
  \end{tabularx}
  \caption{Sample responses when using \texttt{Llama-3.1-8B-Instruct} as the base model with varying numbers of natural language and compressed contexts. `2/8' means using 2 natural language and 8 compressed context. Exact matches with the ground-truth answer is in \textbf{bold} and semantic similar parts are in \gray{gray}. As the number of natural language contexts increase, the model answers are more detailed.}
  \label{tab:sensitivity_answers}
\end{table}

%% file: tab/tab-prompt.tex
\begin{table}[ht]
\centering
\begin{tabular}{p{0.96\textwidth}}
\toprule
\textbf{[Embedding Alignment]} 
\newline
\texttt{<C>} means the same as: \texttt{<Sentence>}
\\
\midrule
\textbf{[Context Reconstruction]}
\newline
Interpret the following tokens as a single document: 
\texttt{<C>} \texttt{<C>} \ldots \texttt{<C>}:  \texttt{<Paragraph>}
\\
\midrule
\textbf{[Instruction-tuning / Inference]}
\newline
Using the context and additional context, answer the following question: \emph{\texttt{<question>}}
\newline
\textbf{Context}: \emph{\texttt{<context>}} 
\newline
\textbf{Additional Context}: 
\newline 
1. \texttt{<C>}, \texttt{<C>}, \ldots,  \texttt{<C>};
\newline 
2. \texttt{<C>}, \texttt{<C>}, \ldots,  \texttt{<C>};
\newline
\textbf{Question}: \emph{\texttt{<Question>}}
\newline
\textbf{Your Answer}: \emph{\texttt{<Answer>}}
\newline
\textbf{Judgment}:  \\
\bottomrule
\end{tabular}
\caption{Prompt for pretraining, instruction-tuning, and inference. \texttt{<C>} indicate positions for the compression vectors}
\label{tab:prompt}
\end{table}

%% file: tab/tab-context_reconstruction.tex
\begin{table*}[ht]
\centering
\begin{tabular}{p{0.44\textwidth}|p{0.48\textwidth}}
\toprule
\textbf{Prediction} & \textbf{Ground-truth} \\
\midrule
\textbf{\# Anti-scam dataset}
\newline 
Collecting \textbf{human-human conversational data} to create a dataset for training and evaluating \textbf{anti-scam} \underline{models}. 
We \textbf{collect conversations} between \textbf{users} and \textbf{attackers who aim to gather customer information} from \textbf{Amazon customer service scam scenarios}. We collected \underline{220} anti-scam conversational data from \textbf{Amazon customers} through a Turkers' platform, which are \textbf{human-human dialogues}. 
The average length of a conversation is \underline{11.5 turns} and the average length is 11 words. \textbf{172 out of 220 users successfully identified attackers}, indicating that \textbf{the attackers are well-trained} in their scam attack strategy. We recruited \textbf{two experienced annotators} to evaluate the quality of the annotated data. 
& 
\textbf{\#\# AntiScam Dataset}\newline
To enrich available \underline{non-collaborative} task datasets, we created a corpus of \textbf{human-human anti-scam dialogs} in order to learn \underline{human elicitation strategies}. 
We chose a popular \textbf{Amazon customer service scam scenario} to collect dialogs between users and attackers who aim to collect users information. We posted a role-playing task on the Amazon Mechanical Turk platform and collected a typing conversation dataset named AntiScam. We collected 220 human-human dialogs. The average conversation length is 12.45 turns and the average utterance length is 11.13 words. \textbf{Only 172 out of 220 users successfully identified their partner as an attacker}, suggesting that \textbf{the attackers are well trained} and not too easily identifiable. We recruited \textbf{two expert annotators} who have linguistic training to \underline{annotate 3,044 sentences in 100 dialogs}, \underline{achieving a 0.874 averaged weighted kappa value}. \\
\midrule
\textbf{Exploration of oil in Nigeria} \textbf{began around 1900}, when oil was discovered in \underline{commercial quantities} in the \underline{Niger Delta region}. However, \textbf{large-quantities} was only \textbf{discovered later} in \textbf{1956} in \textbf{Oloibiri}. & Although the history of \textbf{oil exploration in Nigeria} \textbf{dates back to 1903}, \underline{non-commercial quantities of oil} were not discovered there until 1953. \textbf{Commercial amounts of crude oil} were \textbf{later discovered} in \textbf{Oloibiri, Nigeria} in \textbf{1956}. \\
\midrule
\textbf{The Great Trek} was a series of \textbf{migrations} of \textbf{Dutch-speaking settlers} from \textbf{Cape Colony in South Africa}, which \textbf{began in 1836} and \underline{lasted for several years}. & \textbf{The Great Trek} was an \underline{eastward} \textbf{migration} of \textbf{Dutch-speaking settlers} who travelled by \underline{wagon trains} from \textbf{the Cape Colony} into the interior of modern \textbf{South Africa} \textbf{from 1836 onwards}. The exploratory treks, however, \underline{arrived at the bay of Port Natal} in \underline{February 1835}. \\
\midrule
\textbf{The history of music} is \textbf{the study of music and its development over time}, from \textbf{prehistoric times} to \textbf{the present day}. The oldest known written music is the song ``Hymn to the Sun'' from \underline{the Sumerian civilization}, which is believed to be \textbf{over 3,000 years} old. & \textbf{The history of music} covers \textbf{the historical development and evolution of music} from \textbf{prehistoric times} to \textbf{present day}. The ``\textbf{oldest known song}'' was \underline{written in cuneiform}, dating to \underline{3400 years ago} from \textbf{Ugarit in Syria}. The first piece of unwritten music was made prior to the \underline{Paleolithic age} \underline{3.3 million years ago}. \\
\bottomrule
\end{tabular}
\caption{Reconstruction quality of compression tokens in \method. Source‑aligned spans are shown in \textbf{bold} and errors are \underline{underlined}. \method faithfully reproduces most original semantics with only minor hallucinations. }
\label{tab:context_reconstruction_mistral7b}
\end{table*}

%% file: tab/tab-overall_compress_llm_metrics_2.tex
\begin{table*}[htbp]
\centering
\begin{tabular}{lcccccccc}
\toprule
\multirow{2}{*}{Model} & \multicolumn{4}{c}{NarrativeQA} & \multicolumn{4}{c}{SQuAD-v2} \\
\cmidrule(lr){2-5} \cmidrule(lr){6-9}
& Rele. & Correct. & Sim. & Faith. & Rele. & Correct. & Sim. & Faith. \\
\midrule
ICAE\_Mistral7B & 52.08 & 16.75 & 51.27 & 21.19 & 67.17 & 51.93 & 75.25 & 69.64 \\
LLMLingua       & 84.42 & 37.03 & 79.95 & 39.66 & 86.63 & 70.66 & 89.70 & 75.76 \\
LongLLMLingua   & 84.17 & 34.38 & 76.67 & 30.86 & 83.73 & 67.90 & 87.72 & 73.98 \\
SARA            & 87.87 & 44.09 & 82.26 & 43.83 & 90.66 & 77.21 & 92.16 & 80.12 \\
\midrule
\multirow{2}{*}{Model} & \multicolumn{4}{c}{TriviaQA} & \multicolumn{4}{c}{HotpotQA} \\
\cmidrule(lr){2-5} \cmidrule(lr){6-9}
& Rele. & Correct. & Sim. & Faith. & Rele. & Correct. & Sim. & Faith. \\
\midrule
ICAE\_Mistral7B & 54.70 & 36.48 & 58.21 & 58.05 & 47.81 & 21.59 & 53.19 & 39.37 \\
LLMLingua       & 71.95 & 68.95 & 82.26 & 61.58 & 61.43 & 41.72 & 73.63 & 75.94 \\
LongLLMLingua   & 70.44 & 70.52 & 82.67 & 72.53 & 61.56 & 41.97 & 74.02 & 77.49 \\
SARA            & 88.92 & 70.63 & 88.14 & 76.47 & 83.09 & 55.55 & 86.94 & 80.03 \\
\bottomrule
\end{tabular}
\caption{LLM-based evaluation results across four datasets under context constraint of $512$ tokens. We report Response Relevance (Rele.), Answer Correctness (Correct.), Semantic Similarity (Sim.), and Faithfulness (Faith.) in percentages.}
\label{tab:overall_compress_llm_metrics_2}
\end{table*}

%% file: tab/tab-vary_retrievers.tex
\begin{table}[!ht]
\centering
\begin{tabular}{lcccccc}
\toprule
Retriever & \multicolumn{2}{c}{QASPER} & \multicolumn{2}{c}{NarrativeQA} & \multicolumn{2}{c}{TriviaQA} \\
 & F-1 & ROUGE-L & F-1 & ROUGE-L & F-1 & ROUGE-L \\
\midrule
SFR & 55.44 & 52.93 & 58.03 & 56.39 & 84.13 & 83.61 \\
BGE & 44.47 & 45.24 & 54.05 & 53.98 & 85.41 & 84.58 \\
BM25 & 36.15 & 39.54 & 56.79 & 55.76 & 83.58 & 83.65 \\
\bottomrule
\end{tabular}
\caption{Generalizability across different retrievers.}
\label{tab:vary_retrievers}
\end{table}

%% file: tab/tab-ood_datasets.tex
\begin{table}[ht]
\centering
\begin{tabular}{lcccc}
\toprule
\textbf{QMSum} & \textbf{Relevance} & \textbf{Correctness} & \textbf{Similarity} & \textbf{Faithfulness} \\
\midrule
Mistral7B & 51.82 & 8.97 & 52.90 & 69.39 \\
+\method & 70.37 & 11.17 & 53.51 & 70.68 \\
\midrule
\textbf{MultifieldQA-en} & \textbf{Relevance} & \textbf{Correctness} & \textbf{Similarity} & \textbf{Faithfulness} \\
\midrule
Mistral7B & 42.32 & 21.97 & 42.09 & 31.61 \\
+\method & 90.04 & 22.24 & 45.13 & 32.56 \\
\midrule
\textbf{2WikiMultiHopQA} & \textbf{Relevance} & \textbf{Correctness} & \textbf{Similarity} & \textbf{Faithfulness} \\
\midrule
Mistral7B & 31.50 & 35.69 & 29.91 & 42.82 \\
+\method & 86.53 & 37.87 & 31.58 & 44.13 \\
\bottomrule
\end{tabular}
\caption{Results on out-of-domain datasets. We report Response Relevance (Relevance), Answer Correctness (Correctness), Semantic Similarity (Similarity), and Faithfulness (Faithfulness).}
\label{tab:ood_datasets}
\end{table}

%% file: neurips_2025.bbl
\begin{thebibliography}{70}
\providecommand{\natexlab}[1]{#1}
\providecommand{\url}[1]{\texttt{#1}}
\expandafter\ifx\csname urlstyle\endcsname\relax
  \providecommand{\doi}[1]{doi: #1}\else
  \providecommand{\doi}{doi: \begingroup \urlstyle{rm}\Url}\fi

\bibitem[Asai et~al.(2023)Asai, Wu, Wang, Sil, and Hajishirzi]{asai2023self}
Akari Asai, Zeqiu Wu, Yizhong Wang, Avirup Sil, and Hannaneh Hajishirzi.
\newblock Self-rag: Learning to retrieve, generate, and critique through self-reflection.
\newblock In \emph{ICLR}, 2023.

\bibitem[Bai et~al.(2024)Bai, Lv, Zhang, Lyu, Tang, Huang, Du, Liu, Zeng, Hou, et~al.]{bai2024longbench}
Yushi Bai, Xin Lv, Jiajie Zhang, Hongchang Lyu, Jiankai Tang, Zhidian Huang, Zhengxiao Du, Xiao Liu, Aohan Zeng, Lei Hou, et~al.
\newblock Longbench: A bilingual, multitask benchmark for long context understanding.
\newblock In \emph{ACL}, pages 3119--3137, 2024.

\bibitem[Bengio et~al.(2009)Bengio, Louradour, Collobert, and Weston]{bengio2009curriculum}
Yoshua Bengio, J{\'e}r{\^o}me Louradour, Ronan Collobert, and Jason Weston.
\newblock Curriculum learning.
\newblock In \emph{ICML}, pages 41--48, 2009.

\bibitem[Cheng et~al.(2024)Cheng, Wang, Zhang, Ge, Chen, Wei, Zhang, and Zhao]{cheng2024xrag}
Xin Cheng, Xun Wang, Xingxing Zhang, Tao Ge, Si-Qing Chen, Furu Wei, Huishuai Zhang, and Dongyan Zhao.
\newblock xrag: Extreme context compression for retrieval-augmented generation with one token.
\newblock \emph{arXiv:2405.13792}, 2024.

\bibitem[Chevalier et~al.(2023)Chevalier, Wettig, Ajith, and Chen]{chevalier2023adapting}
Alexis Chevalier, Alexander Wettig, Anirudh Ajith, and Danqi Chen.
\newblock Adapting language models to compress contexts.
\newblock In \emph{EMNLP}, pages 3829--3846, 2023.

\bibitem[Chirkova et~al.(2025)Chirkova, Formal, Nikoulina, and Clinchant]{chirkova2025provence}
Nadezhda Chirkova, Thibault Formal, Vassilina Nikoulina, and St{\'e}phane Clinchant.
\newblock Provence: efficient and robust context pruning for retrieval-augmented generation.
\newblock \emph{arXiv:2501.16214}, 2025.

\bibitem[Dasigi et~al.(2021)Dasigi, Lo, Beltagy, Cohan, Smith, and Gardner]{dasigi2021dataset}
Pradeep Dasigi, Kyle Lo, Iz~Beltagy, Arman Cohan, Noah~A Smith, and Matt Gardner.
\newblock A dataset of information-seeking questions and answers anchored in research papers.
\newblock In \emph{NAACL}, pages 4599--4610, 2021.

\bibitem[Ding et~al.(2023)Ding, Ma, Dong, Zhang, Huang, Wang, Zheng, and Wei]{ding2023longnet}
Jiayu Ding, Shuming Ma, Li~Dong, Xingxing Zhang, Shaohan Huang, Wenhui Wang, Nanning Zheng, and Furu Wei.
\newblock Longnet: Scaling transformers to 1,000,000,000 tokens.
\newblock \emph{arXiv:2307.02486}, 2023.

\bibitem[Edge et~al.(2024)Edge, Trinh, Cheng, Bradley, Chao, Mody, Truitt, and Larson]{edge2024local}
Darren Edge, Ha~Trinh, Newman Cheng, Joshua Bradley, Alex Chao, Apurva Mody, Steven Truitt, and Jonathan Larson.
\newblock From local to global: A graph rag approach to query-focused summarization.
\newblock \emph{arXiv:2404.16130}, 2024.

\bibitem[Es et~al.(2024)Es, James, Anke, and Schockaert]{es2024ragas}
Shahul Es, Jithin James, Luis~Espinosa Anke, and Steven Schockaert.
\newblock Ragas: Automated evaluation of retrieval augmented generation.
\newblock In \emph{EACL}, pages 150--158, 2024.

\bibitem[Gao et~al.(2023)Gao, Yen, Yu, and Chen]{gao2023enabling}
Tianyu Gao, Howard Yen, Jiatong Yu, and Danqi Chen.
\newblock Enabling large language models to generate text with citations.
\newblock In \emph{EMNLP}, pages 6465--6488. ACL, 2023.

\bibitem[Ge et~al.(2024)Ge, Jing, Wang, Wang, Chen, and Wei]{gecontext}
Tao Ge, Hu~Jing, Lei Wang, Xun Wang, Si-Qing Chen, and Furu Wei.
\newblock In-context autoencoder for context compression in a large language model.
\newblock In \emph{ICLR}, 2024.

\bibitem[Ho et~al.(2020)Ho, Nguyen, Sugawara, and Aizawa]{ho2020constructing}
Xanh Ho, Anh-Khoa~Duong Nguyen, Saku Sugawara, and Akiko Aizawa.
\newblock Constructing a multi-hop qa dataset for comprehensive evaluation of reasoning steps.
\newblock In \emph{COLING}, pages 6609--6625, 2020.

\bibitem[Hu et~al.(2021)Hu, Wallis, Allen-Zhu, Li, Wang, Wang, Chen, et~al.]{hu2021lora}
Edward~J Hu, Phillip Wallis, Zeyuan Allen-Zhu, Yuanzhi Li, Shean Wang, Lu~Wang, Weizhu Chen, et~al.
\newblock Lora: Low-rank adaptation of large language models.
\newblock In \emph{ICLR}, 2021.

\bibitem[Hwang et~al.(2024)Hwang, Cho, Jeong, Song, Han, and Park]{hwang2024exit}
Taeho Hwang, Sukmin Cho, Soyeong Jeong, Hoyun Song, SeungYoon Han, and Jong~C Park.
\newblock Exit: Context-aware extractive compression for enhancing retrieval-augmented generation.
\newblock \emph{arXiv:2412.12559}, 2024.

\bibitem[Izacard et~al.(2023)Izacard, Lewis, Lomeli, Hosseini, Petroni, Schick, Dwivedi-Yu, Joulin, Riedel, and Grave]{izacard2023atlas}
Gautier Izacard, Patrick Lewis, Maria Lomeli, Lucas Hosseini, Fabio Petroni, Timo Schick, Jane Dwivedi-Yu, Armand Joulin, Sebastian Riedel, and Edouard Grave.
\newblock Atlas: Few-shot learning with retrieval augmented language models.
\newblock \emph{JMLR}, 24\penalty0 (251):\penalty0 1--43, 2023.

\bibitem[Jiang et~al.(2023{\natexlab{a}})Jiang, Sablayrolles, Mensch, Bamford, Chaplot, Casas, Bressand, Lengyel, Lample, Saulnier, et~al.]{jiang2023mistral}
Albert~Q Jiang, Alexandre Sablayrolles, Arthur Mensch, Chris Bamford, Devendra~Singh Chaplot, Diego de~las Casas, Florian Bressand, Gianna Lengyel, Guillaume Lample, Lucile Saulnier, et~al.
\newblock Mistral 7b.
\newblock \emph{arXiv preprint arXiv:2310.06825}, 2023{\natexlab{a}}.

\bibitem[Jiang et~al.(2023{\natexlab{b}})Jiang, Wu, Lin, Yang, and Qiu]{jiang2023llmlingua}
Huiqiang Jiang, Qianhui Wu, Chin-Yew Lin, Yuqing Yang, and Lili Qiu.
\newblock Llmlingua: Compressing prompts for accelerated inference of large language models.
\newblock In \emph{EMNLP}, pages 13358--13376, 2023{\natexlab{b}}.

\bibitem[Jiang et~al.(2024)Jiang, Wu, Luo, Li, Lin, Yang, and Qiu]{jiang2023longllmlingua}
Huiqiang Jiang, Qianhui Wu, Xufang Luo, Dongsheng Li, Chin-Yew Lin, Yuqing Yang, and Lili Qiu.
\newblock Longllmlingua: Accelerating and enhancing llms in long context scenarios via prompt compression.
\newblock In \emph{ACL}, 2024.

\bibitem[Jin et~al.(2024{\natexlab{a}})Jin, Han, Yang, Jiang, Liu, Chang, Chen, and Hu]{jin2024llm}
Hongye Jin, Xiaotian Han, Jingfeng Yang, Zhimeng Jiang, Zirui Liu, Chia-Yuan Chang, Huiyuan Chen, and Xia Hu.
\newblock Llm maybe longlm: Selfextend llm context window without tuning.
\newblock In \emph{ICML}, pages 22099--22114, 2024{\natexlab{a}}.

\bibitem[Jin et~al.(2024{\natexlab{b}})Jin, Chandra, Verma, Hu, De~Choudhury, and Kumar]{jin2024better}
Yiqiao Jin, Mohit Chandra, Gaurav Verma, Yibo Hu, Munmun De~Choudhury, and Srijan Kumar.
\newblock Better to ask in english: Cross-lingual evaluation of large language models for healthcare queries.
\newblock In \emph{Web Conference}, pages 2627--2638, 2024{\natexlab{b}}.

\bibitem[Joshi et~al.(2017)Joshi, Choi, Weld, and Zettlemoyer]{joshi2017triviaqa}
Mandar Joshi, Eunsol Choi, Daniel~S Weld, and Luke Zettlemoyer.
\newblock Triviaqa: A large scale distantly supervised challenge dataset for reading comprehension.
\newblock In \emph{ACL}, pages 1601--1611, 2017.

\bibitem[Kim et~al.(2024{\natexlab{a}})Kim, Yeom, Yun, and Song]{kimcompressed}
Jang-Hyun Kim, Junyoung Yeom, Sangdoo Yun, and Hyun~Oh Song.
\newblock Compressed context memory for online language model interaction.
\newblock In \emph{ICLR}, 2024{\natexlab{a}}.

\bibitem[Kim et~al.(2024{\natexlab{b}})Kim, Lee, Kwon, Gu, Kim, Cho, Sohn, and Choi]{LinqAIResearch2024}
Junseong Kim, Seolhwa Lee, Jihoon Kwon, Sangmo Gu, Yejin Kim, Minkyung Cho, Jy-yong Sohn, and Chanyeol Choi.
\newblock Linq-embed-mistral:elevating text retrieval with improved gpt data through task-specific control and quality refinement.
\newblock Linq AI Research Blog, 2024{\natexlab{b}}.
\newblock URL \url{https://getlinq.com/blog/linq-embed-mistral/}.

\bibitem[Ko{\v{c}}isk{\`y} et~al.(2018)Ko{\v{c}}isk{\`y}, Schwarz, Blunsom, Dyer, Hermann, Melis, and Grefenstette]{kovcisky2018narrativeqa}
Tom{\'a}{\v{s}} Ko{\v{c}}isk{\`y}, Jonathan Schwarz, Phil Blunsom, Chris Dyer, Karl~Moritz Hermann, G{\'a}bor Melis, and Edward Grefenstette.
\newblock The narrativeqa reading comprehension challenge.
\newblock \emph{Transactions of the Association for Computational Linguistics}, 6:\penalty0 317--328, 2018.

\bibitem[Lewis et~al.(2020)Lewis, Perez, Piktus, Petroni, Karpukhin, Goyal, K{\"u}ttler, Lewis, Yih, Rockt{\"a}schel, et~al.]{lewis2020retrieval}
Patrick Lewis, Ethan Perez, Aleksandra Piktus, Fabio Petroni, Vladimir Karpukhin, Naman Goyal, Heinrich K{\"u}ttler, Mike Lewis, Wen-tau Yih, Tim Rockt{\"a}schel, et~al.
\newblock Retrieval-augmented generation for knowledge-intensive nlp tasks.
\newblock \emph{NeurIPS}, 33:\penalty0 9459--9474, 2020.

\bibitem[Li et~al.(2023{\natexlab{a}})Li, Liu, Xiao, and Shao]{li2023making}
Chaofan Li, Zheng Liu, Shitao Xiao, and Yingxia Shao.
\newblock Making large language models a better foundation for dense retrieval.
\newblock \emph{arXiv:2312.15503}, 2023{\natexlab{a}}.

\bibitem[Li et~al.(2024{\natexlab{a}})Li, He, Guo, Bu, Bai, Liu, Liu, Qu, Li, Ouyang, et~al.]{li2024graphreader}
Shilong Li, Yancheng He, Hangyu Guo, Xingyuan Bu, Ge~Bai, Jie Liu, Jiaheng Liu, Xingwei Qu, Yangguang Li, Wanli Ouyang, et~al.
\newblock Graphreader: Building graph-based agent to enhance long-context abilities of large language models.
\newblock In \emph{EMNLP}, pages 12758--12786, 2024{\natexlab{a}}.

\bibitem[Li et~al.(2024{\natexlab{b}})Li, Li, Ramos, Tang, and Elliott]{li2024understanding}
Wenyan Li, Jiaang Li, Rita Ramos, Raphael Tang, and Desmond Elliott.
\newblock Understanding retrieval robustness for retrieval-augmented image captioning.
\newblock In \emph{ACL}, pages 9285--9299, 2024{\natexlab{b}}.

\bibitem[Li et~al.(2023{\natexlab{b}})Li, Zhang, Zhang, Long, Xie, and Zhang]{li2023towards}
Zehan Li, Xin Zhang, Yanzhao Zhang, Dingkun Long, Pengjun Xie, and Meishan Zhang.
\newblock Towards general text embeddings with multi-stage contrastive learning.
\newblock \emph{arXiv:2308.03281}, 2023{\natexlab{b}}.

\bibitem[Li et~al.(2025)Li, Chen, and Jeon]{li2025grappi}
Ziwen Li, Xiang'Anthony' Chen, and Youngseung Jeon.
\newblock Grappi: A retrieve-divide-solve graphrag framework for large-scale protein-protein interaction exploration.
\newblock In \emph{NAACL}, 2025.

\bibitem[Lin(2004)]{lin2004rouge}
Chin-Yew Lin.
\newblock Rouge: A package for automatic evaluation of summaries.
\newblock In \emph{Text summarization branches out}, pages 74--81, 2004.

\bibitem[Liu et~al.(2023)Liu, Li, Wu, and Lee]{liu2023visual}
Haotian Liu, Chunyuan Li, Qingyang Wu, and Yong~Jae Lee.
\newblock Visual instruction tuning.
\newblock \emph{NeurIPS}, 36:\penalty0 34892--34916, 2023.

\bibitem[Liu(2022)]{Liu_LlamaIndex_2022}
Jerry Liu.
\newblock {LlamaIndex}.
\newblock \url{https://github.com/jerryjliu/llama_index}, 11 2022.
\newblock DOI: \href{https://doi.org/10.5281/zenodo.1234}{10.5281/zenodo.1234}.

\bibitem[Liu et~al.(2025)Liu, Jin, Li, Wong, Wen, Sun, Chen, Xie, and Wang]{liu2025culturevlm}
Shudong Liu, Yiqiao Jin, Cheng Li, Derek~F Wong, Qingsong Wen, Lichao Sun, Haipeng Chen, Xing Xie, and Jindong Wang.
\newblock Culturevlm: Characterizing and improving cultural understanding of vision-language models for over 100 countries.
\newblock \emph{arXiv:2501.01282}, 2025.

\bibitem[Louis et~al.(2025{\natexlab{a}})Louis, D{\'e}jean, and Clinchant]{louis2025pisco}
Maxime Louis, Herv{\'e} D{\'e}jean, and St{\'e}phane Clinchant.
\newblock Pisco: Pretty simple compression for retrieval-augmented generation.
\newblock \emph{arXiv:2501.16075}, 2025{\natexlab{a}}.

\bibitem[Louis et~al.(2025{\natexlab{b}})Louis, Formal, Dejean, and Clinchant]{louis2025oscar}
Maxime Louis, Thibault Formal, Herv{\'e} Dejean, and St{\'e}phane Clinchant.
\newblock Oscar: Online soft compression and reranking.
\newblock \emph{arXiv:2504.07109}, 2025{\natexlab{b}}.

\bibitem[Luo et~al.(2025)Luo, Luo, Chen, Xiao, Ju, and Zhang]{luo2025semi}
Junyu Luo, Xiao Luo, Xiusi Chen, Zhiping Xiao, Wei Ju, and Ming Zhang.
\newblock Semi-supervised fine-tuning for large language models.
\newblock In \emph{NAACL}, pages 2795--2808, 2025.

\bibitem[Meng et~al.(2024)Meng, Liu, Joty, Xiong, Zhou, and Yavuz]{SFRAIResearch2024}
Rui Meng, Ye~Liu, Shafiq~Rayhan Joty, Caiming Xiong, Yingbo Zhou, and Semih Yavuz.
\newblock Sfr-embedding-mistral:enhance text retrieval with transfer learning.
\newblock Salesforce AI Research Blog, 2024.
\newblock URL \url{https://www.salesforce.com/blog/sfr-embedding/}.

\bibitem[Mu et~al.(2023)Mu, Li, and Goodman]{mu2023learning}
Jesse Mu, Xiang Li, and Noah Goodman.
\newblock Learning to compress prompts with gist tokens.
\newblock \emph{NeurIPS}, 36:\penalty0 19327--19352, 2023.

\bibitem[Muennighoff et~al.(2023)Muennighoff, Tazi, Magne, and Reimers]{muennighoff2023mteb}
Niklas Muennighoff, Nouamane Tazi, Loic Magne, and Nils Reimers.
\newblock Mteb: Massive text embedding benchmark.
\newblock In \emph{ACL}, pages 2014--2037, 2023.

\bibitem[Munkhdalai et~al.(2024)Munkhdalai, Faruqui, and Gopal]{munkhdalai2024leave}
Tsendsuren Munkhdalai, Manaal Faruqui, and Siddharth Gopal.
\newblock Leave no context behind: Efficient infinite context transformers with infini-attention.
\newblock \emph{arXiv:2404.07143}, 2024.

\bibitem[OpenAI(2025)]{gpt4o}
OpenAI.
\newblock Gpt-4o, 2025.
\newblock URL \url{https://chat.openai.com/}.

\bibitem[Pan et~al.(2024)Pan, Wu, Jiang, Xia, Luo, Zhang, Lin, R{\"u}hle, Yang, Lin, et~al.]{pan2024llmlingua}
Zhuoshi Pan, Qianhui Wu, Huiqiang Jiang, Menglin Xia, Xufang Luo, Jue Zhang, Qingwei Lin, Victor R{\"u}hle, Yuqing Yang, Chin-Yew Lin, et~al.
\newblock Llmlingua-2: Data distillation for efficient and faithful task-agnostic prompt compression.
\newblock In \emph{ACL}, pages 963--981, 2024.

\bibitem[Pang et~al.(2022)Pang, Parrish, Joshi, Nangia, Phang, Chen, Padmakumar, Ma, Thompson, He, et~al.]{pang2022quality}
Richard~Yuanzhe Pang, Alicia Parrish, Nitish Joshi, Nikita Nangia, Jason Phang, Angelica Chen, Vishakh Padmakumar, Johnny Ma, Jana Thompson, He~He, et~al.
\newblock Quality: Question answering with long input texts, yes!
\newblock In \emph{NAACL}, pages 5336--5358, 2022.

\bibitem[Paszke et~al.(2019)Paszke, Gross, Massa, Lerer, Bradbury, Chanan, Killeen, Lin, Gimelshein, Antiga, et~al.]{paszke2019pytorch}
Adam Paszke, Sam Gross, Francisco Massa, Adam Lerer, James Bradbury, Gregory Chanan, Trevor Killeen, Zeming Lin, Natalia Gimelshein, Luca Antiga, et~al.
\newblock Pytorch: An imperative style, high-performance deep learning library.
\newblock In \emph{NeurIPS}, volume~32, 2019.

\bibitem[Rajpurkar et~al.(2018)Rajpurkar, Jia, and Liang]{rajpurkar2018know}
Pranav Rajpurkar, Robin Jia, and Percy Liang.
\newblock Know what you don’t know: Unanswerable questions for squad.
\newblock In \emph{ACL}, pages 784--789, 2018.

\bibitem[Reimers and Gurevych(2019)]{reimers-2019-sentence-bert}
Nils Reimers and Iryna Gurevych.
\newblock Sentence-bert: Sentence embeddings using siamese bert-networks.
\newblock In \emph{EMNLP}. Association for Computational Linguistics, 11 2019.
\newblock URL \url{https://arxiv.org/abs/1908.10084}.

\bibitem[Risch et~al.(2021)Risch, M{\"o}ller, Gutsch, and Pietsch]{risch2021semantic}
Julian Risch, Timo M{\"o}ller, Julian Gutsch, and Malte Pietsch.
\newblock Semantic answer similarity for evaluating question answering models.
\newblock In \emph{Proceedings of the 3rd Workshop on Machine Reading for Question Answering}, pages 149--157, 2021.

\bibitem[Robertson et~al.(2004)Robertson, Zaragoza, and Taylor]{robertson2004simple}
Stephen Robertson, Hugo Zaragoza, and Michael Taylor.
\newblock Simple bm25 extension to multiple weighted fields.
\newblock In \emph{CIKM}, pages 42--49, 2004.

\bibitem[Sarthi et~al.(2024)Sarthi, Abdullah, Tuli, Khanna, Goldie, and Manning]{sarthi2024raptor}
Parth Sarthi, Salman Abdullah, Aditi Tuli, Shubh Khanna, Anna Goldie, and Christopher~D. Manning.
\newblock Raptor: Recursive abstractive processing for tree-organized retrieval.
\newblock In \emph{ICLR}, 2024.

\bibitem[Shannon(1948)]{shannon1948mathematical}
Claude~E Shannon.
\newblock A mathematical theory of communication.
\newblock \emph{The Bell system technical journal}, 27\penalty0 (3):\penalty0 379--423, 1948.

\bibitem[Sharma et~al.(2024)Sharma, Kumar, and Li]{sharma2024og}
Kartik Sharma, Peeyush Kumar, and Yunqing Li.
\newblock Og-rag: Ontology-grounded retrieval-augmented generation for large language models.
\newblock \emph{arXiv:2412.15235}, 2024.

\bibitem[Song et~al.(2020)Song, Tan, Qin, Lu, and Liu]{song2020mpnet}
Kaitao Song, Xu~Tan, Tao Qin, Jianfeng Lu, and Tie-Yan Liu.
\newblock Mpnet: Masked and permuted pre-training for language understanding.
\newblock \emph{NeurIPS}, 33:\penalty0 16857--16867, 2020.

\bibitem[Wang et~al.(2024)Wang, Wang, Gao, Zhang, Wu, Xu, Shi, Wang, Li, Qian, et~al.]{wangsearching}
Xiaohua Wang, Zhenghua Wang, Xuan Gao, Feiran Zhang, Yixin Wu, Zhibo Xu, Tianyuan Shi, Zhengyuan Wang, Shizheng Li, Qi~Qian, et~al.
\newblock Searching for best practices in retrieval-augmented generation.
\newblock In \emph{EMNLP}, 2024.

\bibitem[Wang et~al.(2021)Wang, Chen, and Zhu]{wang2021survey}
Xin Wang, Yudong Chen, and Wenwu Zhu.
\newblock A survey on curriculum learning.
\newblock \emph{IEEE transactions on pattern analysis and machine intelligence}, 44\penalty0 (9):\penalty0 4555--4576, 2021.

\bibitem[Wei et~al.(2025)Wei, Chen, and Meng]{wei2024instructrag}
Zhepei Wei, Wei-Lin Chen, and Yu~Meng.
\newblock Instructrag: Instructing retrieval-augmented generation via self-synthesized rationales.
\newblock In \emph{ICLR}, 2025.

\bibitem[Wolf et~al.(2020)Wolf, Debut, Sanh, Chaumond, Delangue, Moi, Cistac, Rault, Louf, Funtowicz, et~al.]{wolf2020transformers}
Thomas Wolf, Lysandre Debut, Victor Sanh, Julien Chaumond, Clement Delangue, Anthony Moi, Pierric Cistac, Tim Rault, R{\'e}mi Louf, Morgan Funtowicz, et~al.
\newblock Transformers: State-of-the-art natural language processing.
\newblock In \emph{EMNLP}, pages 38--45, 2020.

\bibitem[Xiao et~al.(2024{\natexlab{a}})Xiao, Jin, Bai, Wu, Yang, Luo, Yu, Zhao, Liu, Chen, et~al.]{xiao2023large}
Yijia Xiao, Yiqiao Jin, Yushi Bai, Yue Wu, Xianjun Yang, Xiao Luo, Wenchao Yu, Xujiang Zhao, Yanchi Liu, Haifeng Chen, et~al.
\newblock Large language models can be good privacy protection learners.
\newblock In \emph{EMNLP}, 2024{\natexlab{a}}.

\bibitem[Xiao et~al.(2024{\natexlab{b}})Xiao, Sun, Jin, Wang, and Wang]{xiao2024proteingpt}
Yijia Xiao, Edward Sun, Yiqiao Jin, Qifan Wang, and Wei Wang.
\newblock Proteingpt: Multimodal llm for protein property prediction and structure understanding.
\newblock \emph{arXiv preprint arXiv:2408.11363}, 2024{\natexlab{b}}.

\bibitem[Xu et~al.(2024)Xu, Shi, and Choi]{xu2024recomp}
Fangyuan Xu, Weijia Shi, and Eunsol Choi.
\newblock Recomp: Improving retrieval-augmented lms with compression and selective augmentation.
\newblock In \emph{ICLR}, 2024.

\bibitem[Yang et~al.(2018)Yang, Qi, Zhang, Bengio, Cohen, Salakhutdinov, and Manning]{yang2018hotpotqa}
Zhilin Yang, Peng Qi, Saizheng Zhang, Yoshua Bengio, William Cohen, Ruslan Salakhutdinov, and Christopher~D Manning.
\newblock Hotpotqa: A dataset for diverse, explainable multi-hop question answering.
\newblock In \emph{EMNLP}, pages 2369--2380, 2018.

\bibitem[Yoon et~al.(2024)Yoon, Lee, Hwang, Jeong, and Kang]{yoon2024compact}
Chanwoong Yoon, Taewhoo Lee, Hyeon Hwang, Minbyul Jeong, and Jaewoo Kang.
\newblock Compact: Compressing retrieved documents actively for question answering.
\newblock In \emph{EMNLP}, 2024.

\bibitem[Yu et~al.(2024)Yu, Ping, Liu, Wang, You, Zhang, Shoeybi, and Catanzaro]{yurankrag}
Yue Yu, Wei Ping, Zihan Liu, Boxin Wang, Jiaxuan You, Chao Zhang, Mohammad Shoeybi, and Bryan Catanzaro.
\newblock Rankrag: Unifying context ranking with retrieval-augmented generation in llms.
\newblock In \emph{NeurIPS}, 2024.

\bibitem[Zhang et~al.(2024{\natexlab{a}})Zhang, Li, Zeng, and Wang]{zhang2024jasper}
Dun Zhang, Jiacheng Li, Ziyang Zeng, and Fulong Wang.
\newblock Jasper and stella: distillation of sota embedding models.
\newblock \emph{arXiv:2412.19048}, 2024{\natexlab{a}}.

\bibitem[Zhang et~al.(2024{\natexlab{b}})Zhang, Zhang, Pang, Zheng, and Zheng]{zhang2024adacomp}
Qianchi Zhang, Hainan Zhang, Liang Pang, Hongwei Zheng, and Zhiming Zheng.
\newblock Adacomp: Extractive context compression with adaptive predictor for retrieval-augmented large language models.
\newblock \emph{arXiv:2409.01579}, 2024{\natexlab{b}}.

\bibitem[Zhang et~al.(2024{\natexlab{c}})Zhang, Chen, Liu, Yao, Ruwase, Chen, Wu, and Wang]{zhangfound}
Zhenyu Zhang, Runjin Chen, Shiwei Liu, Zhewei Yao, Olatunji Ruwase, Beidi Chen, Xiaoxia Wu, and Zhangyang Wang.
\newblock Found in the middle: How language models use long contexts better via plug-and-play positional encoding.
\newblock In \emph{NeurIPS}, 2024{\natexlab{c}}.

\bibitem[Zhao et~al.(2024)Zhao, Wang, Zhang, Jin, Zhu, Chen, and Xie]{zhao2023competeai}
Qinlin Zhao, Jindong Wang, Yixuan Zhang, Yiqiao Jin, Kaijie Zhu, Hao Chen, and Xing Xie.
\newblock Competeai: Understanding the competition behaviors in large language model-based agents.
\newblock In \emph{ICML}, 2024.

\bibitem[Zhong et~al.(2021)Zhong, Yin, Yu, Zaidi, Mutuma, Jha, Hassan, Celikyilmaz, Liu, Qiu, et~al.]{zhong2021qmsum}
Ming Zhong, Da~Yin, Tao Yu, Ahmad Zaidi, Mutethia Mutuma, Rahul Jha, Ahmed Hassan, Asli Celikyilmaz, Yang Liu, Xipeng Qiu, et~al.
\newblock Qmsum: A new benchmark for query-based multi-domain meeting summarization.
\newblock In \emph{NAACL}, pages 5905--5921, 2021.

\bibitem[Zhu et~al.(2025)Zhu, Chen, Hu, Huang, Yuan, Chen, and Birch]{zhu2025generalizing}
Wenhao Zhu, Pinzhen Chen, Hanxu Hu, Shujian Huang, Fei Yuan, Jiajun Chen, and Alexandra Birch.
\newblock Generalizing from short to long: Effective data synthesis for long-context instruction tuning.
\newblock \emph{arXiv:2502.15592}, 2025.

\end{thebibliography}
